\documentclass{article}

\usepackage{microtype}
\usepackage{graphicx}
\usepackage{booktabs} 

\usepackage{multirow} 

\usepackage{hyperref}

\usepackage{amsmath, amsthm}
\usepackage{amsfonts}
\usepackage{amssymb}
\usepackage{mathtools}
\usepackage{nicefrac}

\usepackage{graphicx}
\graphicspath{ {graphs/} }
\usepackage{caption}
\usepackage{subcaption}
\usepackage{cleveref} 

\usepackage{enumitem}
\setlist{noitemsep,topsep=0pt,parsep=0pt,partopsep=0pt,leftmargin=*}

\DeclareMathOperator{\RMSE}{RMSE}
\DeclareMathOperator{\GAP}{G}
\DeclareMathOperator{\E}{\mathbb{E}}

\DeclarePairedDelimiter\norm\lVert\rVert

\usepackage[accepted]{mlsys2020}

\mlsystitlerunning{Taming Momentum in a Distributed Asynchronous Environment}

\begin{document}

\twocolumn[
\mlsystitle{Taming Momentum in a Distributed Asynchronous Environment}



\mlsyssetsymbol{equal}{*}

\begin{mlsysauthorlist}
\mlsysauthor{Ido Hakimi}{equal,tech}
\mlsysauthor{Saar Barkai}{equal,tech}
\mlsysauthor{Moshe Gabel}{to}
\mlsysauthor{Assaf Schuster}{tech}
\end{mlsysauthorlist}

\mlsysaffiliation{to}{Department of Computer and Mathematical Sciences, University of Toronto Scarborough, Toronto, Canada}
\mlsysaffiliation{tech}{Department of Computer Science, Technion -- Israel Institute of Technology, Haifa, Israel}

\mlsyscorrespondingauthor{Ido Hakimi}{idohakimi@gmail.com}
\mlsyscorrespondingauthor{Saar Barkai}{saarbarkai@gmail.com}
\mlsyscorrespondingauthor{Moshe Gabel}{mgabel@cs.toronto.edu}
\mlsyscorrespondingauthor{Assaf Schuster}{assaf@cs.technion.ac.il}

\mlsyskeywords{Distributed, Asynchronous}

\vskip 0.3in

\begin{abstract}
Although distributed computing can significantly reduce the training time of deep neural networks, scaling the training process while maintaining high efficiency and final accuracy is challenging. Distributed asynchronous training enjoys near-linear speedup, but asynchrony causes gradient staleness - the main difficulty in scaling stochastic gradient descent to large clusters. Momentum, which is often used to accelerate convergence and escape local minima, exacerbates the gradient staleness, thereby hindering convergence. We propose DANA: a novel technique for asynchronous distributed SGD with momentum that mitigates gradient staleness by computing the gradient on an estimated future position of the model's parameters. Thereby, we show for the first time that momentum can be fully incorporated in asynchronous training with almost no ramifications to final accuracy. Our evaluation on the CIFAR and ImageNet datasets shows that DANA outperforms existing methods, in both final accuracy and convergence speed while scaling up to a total batch size of $16$K on $64$ asynchronous workers.
\end{abstract}
]



\printAffiliationsAndNotice{\mlsysEqualContribution} 

\section{Introduction}
Modern deep neural networks are comprised of millions of parameters, which require massive amounts of data and long training time \citep{alpha_go,raffel2019exploring}. The steady growth of neural networks over the years has made it impractical to train them from scratch on a single \emph{worker} (computational device). Distributing the computations over several workers can drastically reduce the training time \citep{downpour,mlsys2019_130,mlsys2019_71}. Unfortunately, stochastic gradient descent (SGD), which is typically used to train these networks, is an inherently sequential algorithm. Thus, it is difficult to train deep neural networks on multiple workers, especially when trying to maintain fast convergence and high final test accuracy \citep{keskar2016large,goyal2017accurate,You2020Large}.

\emph{Synchronous SGD} (SSGD) is a straightforward method to distribute the training process across multiple workers: each worker computes the gradient over its own separate mini-batches, which are then aggregated to update a single model. Because SSGD relies on synchronizations to coordinate the workers, its progress is limited by the slowest worker.

\emph{Asynchronous SGD} (ASGD) addresses the drawbacks of SSGD by eliminating synchronization between the workers, allowing it to scale almost linearly. However, eliminating the synchronizations induces \emph{gradient staleness}: gradients were computed on parameters that are older than the parameter server's current parameters. Gradient staleness is one of the main difficulties in scaling ASGD, since it worsens as the number of workers grows \citep{staleness}. Due to gradient staleness, ASGD suffers from slow convergence and reduced final accuracy \citep{chen2016revisiting,cui2016geeps,ben2019demystifying}. In fact, ASGD might not converge at all if the number of workers is too large.

Momentum \citep{momentum} has been demonstrated to accelerate SGD convergence and reduce oscillation \citep{sutskever}. Momentum is crucial for high accuracy and is typically used for training deep neural networks \citep{resnet,wide_resnet}. However, when paired with ASGD, momentum exacerbates the gradient staleness \citep{begets,dai2018toward}, to the point that it diverges when trained on large clusters.

\textbf{Our contribution: } We propose DANA: a novel technique for asynchronous distributed SGD with momentum. By adapting Nesterov's Accelerated Gradient to a distributed setting, DANA computes the gradient on an estimated future position of the model's parameters, thereby mitigating the gradient staleness. As a result, DANA efficiently scales to large clusters, despite using momentum, while maintaining high accuracy and fast convergence. We show for the first time that momentum can be fully incorporated in asynchronous training with almost no ramifications to final accuracy. We evaluate DANA in simulations as well as in two real-world settings: private dedicated compute clusters and cloud data-centers. DANA consistently outperformed other ASGD methods, in both final accuracy and convergence speed while scaling up to a total batch size of $16$K on $64$ asynchronous workers. 

\section{Background}
\label{sec:background}
The goal of SGD is to find the global minimum of $J(\theta)$ where $J$ is the objective function (i.e., loss) and the vector $\theta \in R^k$ is the model's parameters from dimensional $k$. $x_t$ denotes the value of some variable $x$ at iteration $t$. $\xi \in \Xi$ denotes a random variable from $\Xi$, the indices of the entire set of training samples $\Xi = \{1, \cdots, M\}$. $J(\theta; \xi)$ is the stochastic loss function with respect to the training sample indexed by $\xi$. The SGD iterative update rule is the following:
\begin{equation} \label{eq:sgd}
\begin{split}
    g_t = \nabla J(\theta_t; \xi) \quad ; \quad
    \theta_{t+1}=\theta_t-\eta g_t
\end{split}
\end{equation}
Where $\eta$ is the learning rate. We also denote $\nabla J(\theta)$ as the full-batch gradient at point $\theta$:
$\nabla J(\theta) = \frac{1}{M}\sum_{i=1}^M\nabla J(\theta; i)$.

\paragraph{Momentum}
\citet{momentum} proposed momentum, which has been demonstrated to accelerate SGD convergence and reduce oscillation \citep{sutskever}. Momentum can be described as a heavy ball that rolls downhill while accumulating speed on its way towards the minima. The gathered inertia accelerates and smoothes the descent, which helps dampen oscillations and overcome narrow valleys, small humps and local minima \citep{goh2017why}. Mathematically, the momentum update rule (without dampening) is simply an exponentially-weighted moving average of gradients that adds a fraction of the previous momentum vector $v_{t-1}$ to the current momentum vector $v_t$.
\begin{equation} \label{eq:momentum}
\begin{split}
    g_t = \nabla J(\theta_t; \xi) \quad &; \quad 
    v_{t}=\gamma v_{t-1}+g_{t} \\
    \theta_{t+1}&=\theta_t-\eta v_{t}
\end{split}
\end{equation}
The momentum coefficient $\gamma$ in \Cref{eq:momentum} controls the portion of the past gradients that is added to the current momentum vector $v_{t}$. When successive gradients have a similar direction, momentum results in larger steps (higher speed), yielding up to quadratic speedup in the convergence rate for SGD \citep{momentum_loizou,linear_loizou}.

\paragraph{Nesterov}
In the analogy of the heavy ball rolling downhill, a higher speed may cause the heavy ball to overshoot the bottom of the valley (the minima), if it does not slow down in time. \citet{nesterov} proposed \emph{Nesterov's Accelerated Gradient} (NAG), which allows the ball to slow down in advance. NAG approximates $\hat{\theta}_t$, the future value of $\theta_t$, based on the previous momentum vector $v_t$:
\begin{equation} \label{eq:look_ahead}
\begin{split}
    \hat{\theta}_t = \theta_t-\eta \gamma v_{t-1} \quad ;& \quad
    g_t = \nabla J(\hat{\theta}_t; \xi) \\
    v_{t}=\gamma v_{t-1}+g_{t} \quad ;& \quad
    \theta_{t+1}=\theta_t-\eta v_{t}
\end{split}
\end{equation}
NAG computes the gradient using the parameters' estimated future value $\hat{\theta}$ instead of their current value $\theta$. Thus, NAG slows the heavy ball down near the minima so it doesn't overshoot the goal and climb back up the hill. We refer to this attribute as \emph{look-ahead}, since it allows peeking at the future position of $\theta$. The gradient $g_t$ is computed based on the approximated future parameters $\hat{\theta}_t$ and applied to the original parameters $\theta_t$ via $v_t$.
\begin{equation} \label{eq:nag_look_ahead}
\begin{split}
    \theta_{t+1}-\hat{\theta}_t&= \theta_t-\eta v_{t+1} - \theta_t+\eta \gamma v_t \\
    &= \eta \gamma v_t - \eta (\gamma v_t+g_t) 
    = -\eta g_t
\end{split}
\end{equation}
\Cref{eq:nag_look_ahead} shows that the difference between the updated parameters $\theta_{t+1}$ and the approximated future position $\hat{\theta}_t$ is only affected by the newly computed gradient $g_{t}$, and not by $v_t$. Therefore, NAG can accurately compute the gradient even when the momentum vector $v_t$ is large.

\section{Gradient Staleness}
\label{sec:staleness}
\begin{figure*}[t]
    \centering
    \includegraphics[width=\textwidth]{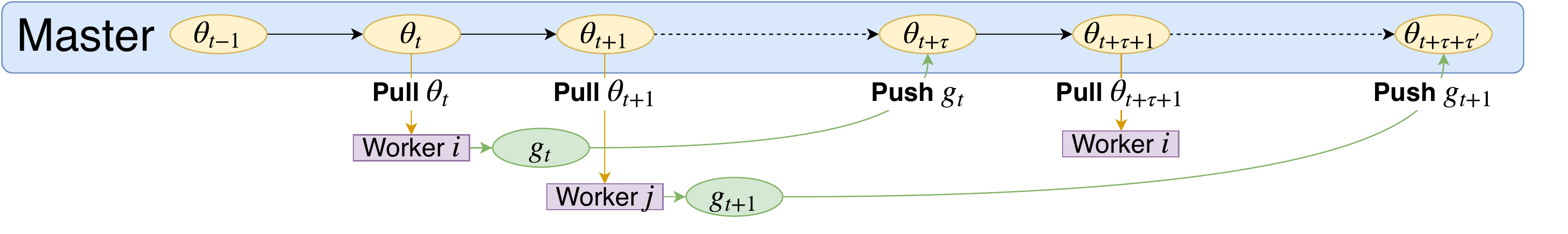}
	\caption{Gradient staleness in the ASGD training process, adapted from \citet{dcasgd}. Gradient $g_t$ is stale, since it is computed from parameters $\theta_t$ but applied to $\theta_{t+\tau}$.}
	\label{fig:flow_asgd}
\end{figure*}
In this work we consider the common implementation of distributed ASGD that uses a parameter server (also referred to as \emph{master}), which is commonly used in large-scale distributed settings \citep{li2014scaling,peng2019generic,mlsys2019_75,mlsys2019_199,mlsys2020_173}. \Cref{fig:flow_asgd} illustrates the ASGD training process and the origin of gradient staleness \citep{recht2011hogwild}. In ASGD training, each worker pulls the up-to-date parameters $\theta_t$ from the master and computes a gradient of a single sample (\Cref{alg:asgd_worker}). Once the computations finish, the worker sends the gradient $g_t$ to the master. The master (\Cref{alg:asgd_master}) then applies the gradient $g_t$ to its current set of parameters $\theta_{t+\tau}$, where $\tau$ is the \emph{lag}. The variable $x$ for worker $i$ is denoted as $x^i$ (for the master, $i=0$). The \emph{lag} $\tau$ of gradient $g_t^i$ is defined as the number of updates the master received from other workers while worker $i$ was computing $g_t^i$.

 \begin{algorithm}[h]
\caption{ASGD: worker}
\label{alg:asgd_worker}
\begin{algorithmic}
    \setlength{\itemindent}{-0.5em} 
    \STATE Receive parameters $\theta_t$ from master
	\STATE Compute gradient $g_t \gets \nabla J(\theta_t; \xi)$
	\STATE Send $g_t$ to master at iteration $t+\tau$
\end{algorithmic}
\end{algorithm}
\begin{algorithm}[h]
\caption{ASGD: master}
\label{alg:asgd_master}
\begin{algorithmic}
    \setlength{\itemindent}{-0.5em} 
    \STATE Receive gradient $g_t$ from worker $i$ (at iteration $t+\tau$)
	\STATE Update master’s weights $\theta_{t+\tau+1} \gets \theta_{t+\tau}-\eta g_t$
	\STATE Send parameters $\theta_{t+\tau+1}$ to worker $i$
\end{algorithmic}
\end{algorithm}
In other words, gradient $g_t^i$ is \emph{stale} if it was computed on parameters $\theta_t$ but applied to $\theta_{t+\tau}$. This gradient staleness is a major obstacle when scaling ASGD: the \emph{lag} $\tau$ increases as the number of workers $N$ grows \citep{staleness}, decreasing gradient accuracy and ultimately reducing the accuracy of the trained model. As a result, ASGD suffers from slow convergence and reduced final accuracy. In fact, ASGD may not converge at all if the number of workers is too large \citep{chen2016revisiting,cui2016geeps}.

\begin{figure*}[t]
\centering
    \begin{subfigure}[t]{0.49\textwidth}
            \includegraphics[width=\textwidth]{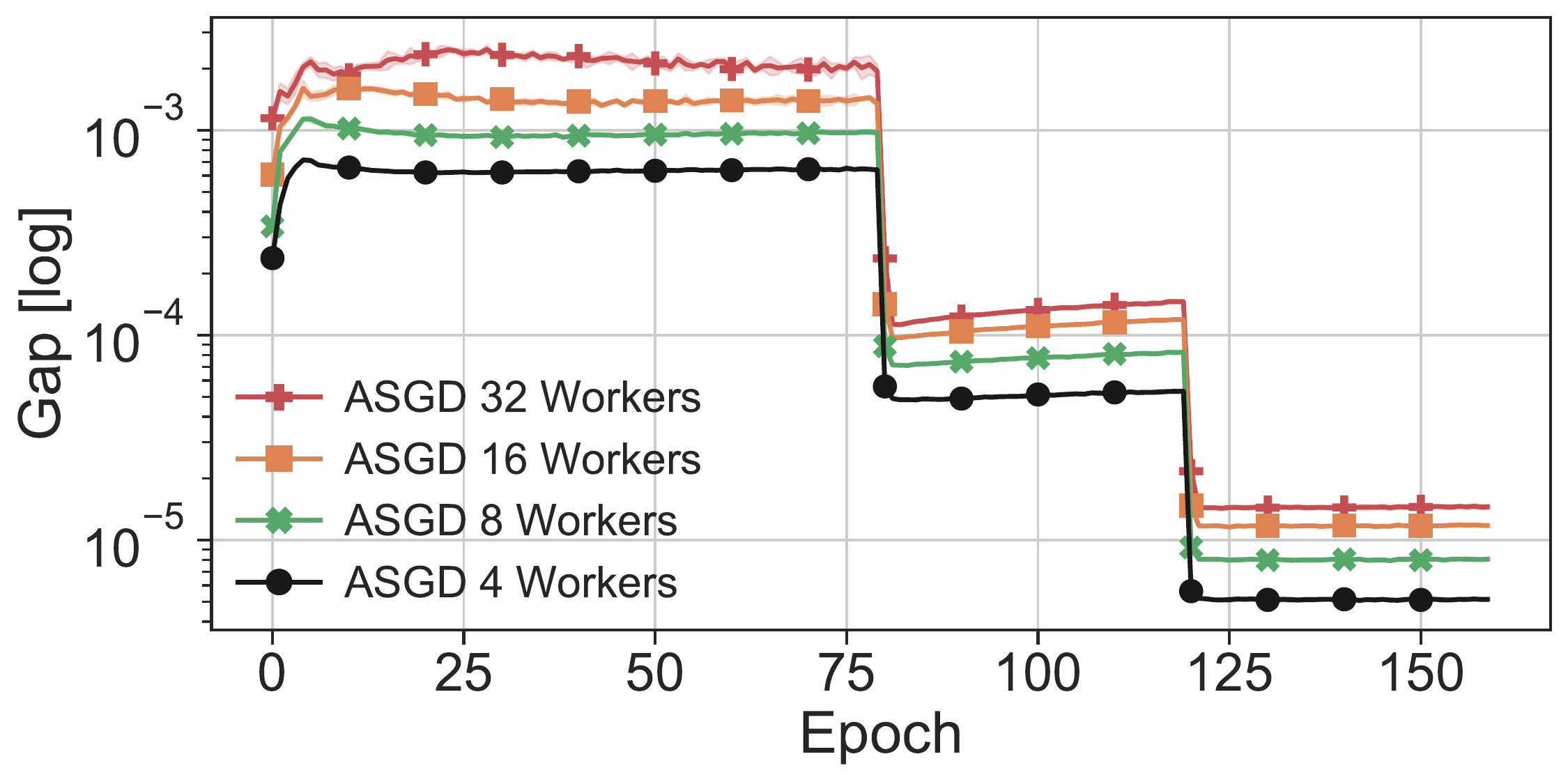}
        	\caption{Comparison of the number of workers.}
        	\label{fig:rmse_workers}
    \end{subfigure}
    \hfill
    \begin{subfigure}[t]{0.49\textwidth}
        	\includegraphics[width=\textwidth]{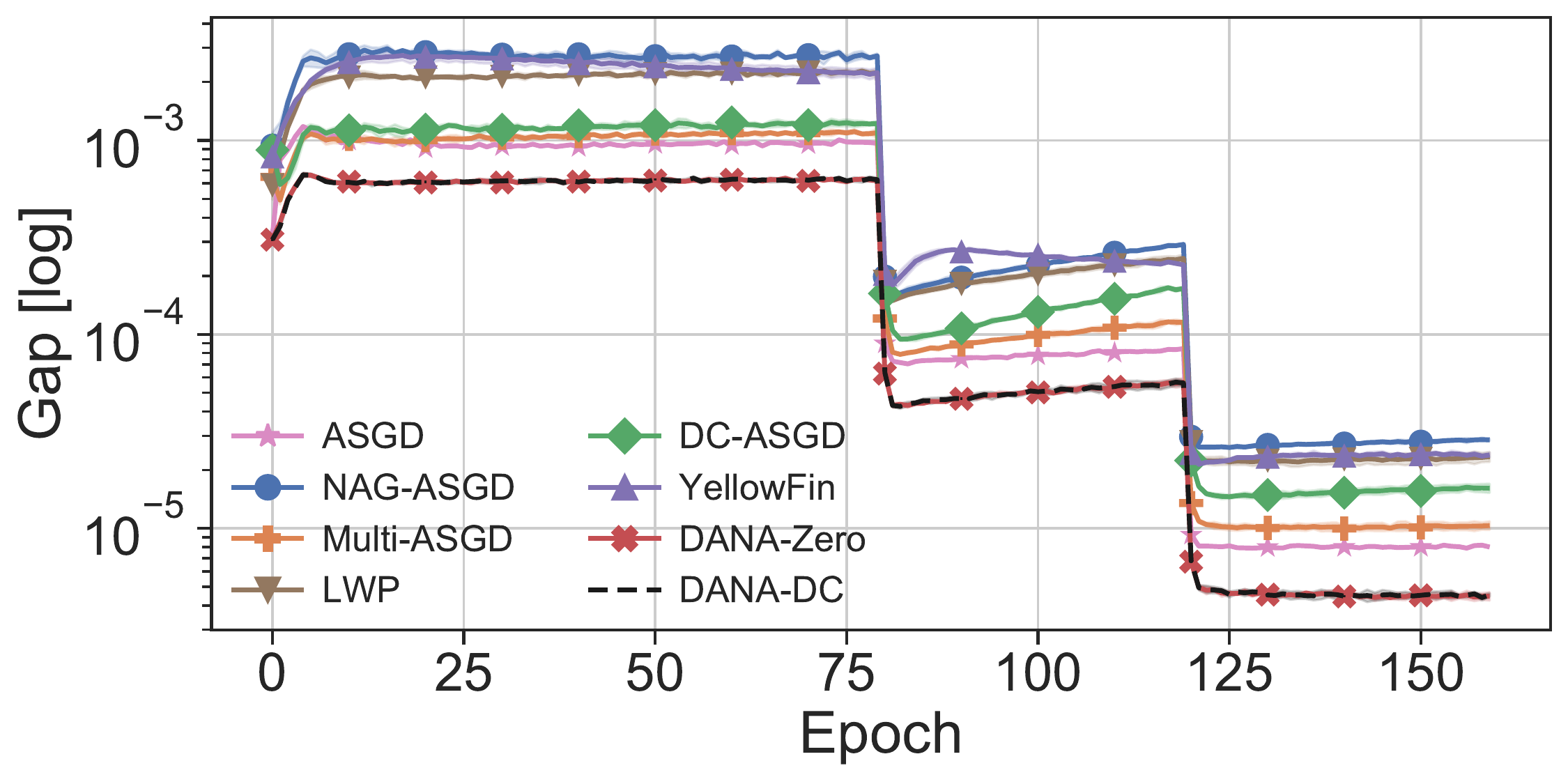}
        	\caption{Algorithm comparison (all with 8 workers).}
        	\label{fig:rmse_algos}
    \end{subfigure}
  \caption{The \emph{gap} between $\theta_{t+\tau}$ and $\theta_{t}$ while training ResNet-20 on the CIFAR-10 dataset with \subref{fig:rmse_workers} different numbers of workers, and \subref{fig:rmse_algos} different asynchronous algorithms. all algorithms share the same worker update schedules and therefore have an identical lag. Adding workers or using momentum increases the effect of the \emph{lag} $\tau$ on the gap. The large drops in the \emph{gap} are caused by learning rate decay. The \emph{gap} drops at the exact same rate at which the learning rate decays, which empirically shows that the \emph{gap} correlates linearly with the learning rate $\eta$. The details for the different algorithms are explained in \Cref{sec:experiments}.}
  \label{fig:RMSE}
\end{figure*}

\paragraph{From Lag to Gap}
Previous works analyze ASGD staleness using the \emph{lag} $\tau$ \citep{staleness,dai2018toward}. We argue that $\tau$ fails to accurately reflect the effects of the staleness. Instead, we measure the gradient staleness with the recently introduced \emph{gap} \citep{Barkai2020Gap-Aware}. We denote $\Delta_{t+\tau} = \theta_{t+\tau} - \theta_{t}$ as the difference between the master and worker parameters, and define the \emph{gap} as:
\begin{equation*}
    \GAP( \Delta_{t+\tau} ) = \RMSE(\Delta_{t+\tau})= \frac{\Vert \Delta_{t+\tau} \Vert_2}{\sqrt{k}}
\end{equation*}
Where $k$ is the number of parameters. When $\Delta_{t+\tau} = 0$, the gradient is computed on the same parameters to which it will be applied. This is the case for sequential and synchronous methods such as SGD and SSGD. However, in asynchronous algorithms more workers result in an increased \emph{lag} $\tau$ and thus a larger \emph{gap}. \Cref{fig:rmse_workers} illustrates that in ASGD adding more worker increases the \emph{gap}. 

Next, we show that the \emph{gap} is directly correlated with gradient accuracy. A common assumption is that the gradient of $J$ is an L-Lipschitz continuous function:
\begin{equation} \label{as:lipschitz}
    \norm{\nabla J\left(x\right) - \nabla J\left(y\right)}_2 \leq L\norm{x-y}_2 \quad, x, y \in \mathbb{R}^k
\end{equation}
Setting $x=\theta_{t+\tau}, y=\theta_t$ into \Cref{as:lipschitz}, we get that the inaccuracy of the stale gradient (with respect to the gradient on $\theta_{t+\tau}$) is bounded by the \emph{gap}:
\begin{equation} \label{eq:lipschitz_gap}
\begin{split}
    \norm{\nabla J\left(\theta_{t+\tau}\right) - \nabla J\left(\theta_{t}\right)}_2 &\leq  L\norm{\theta_{t+\tau}-\theta_{t}}_2 \\
    &=L\cdot\sqrt{k}\cdot\GAP(\Delta_{t+\tau})
\end{split}
\end{equation}
\Cref{eq:lipschitz_gap} shows that a smaller \emph{gap} implies that the stale gradient is more accurate. Conversely, a larger \emph{gap} means a larger upper bound on the inaccuracy of the stale gradient. The advantage of measuring the staleness using the \emph{gap} instead of the \emph{lag} can be illustrated by a simple extreme example of a worker with a \emph{lag} of $\tau=2$ at time step $t+\tau$. If the two previous updates are in exactly opposite directions and are of the same magnitude, the worker will compute the gradient on the same parameters as the master $\theta_{t+\tau}=\theta_{t}$. Therefore, the gradient will be accurately computed $\nabla J\left(\theta_{t+\tau}\right) = \nabla J\left(\theta_{t}\right)$, as if the \emph{lag} were zero. However, the \emph{lag} remains the same ($\tau=2$), while the \emph{gap} adjusts according to the two updates and is indeed equal to zero.

\paragraph{The Effect of Momentum}
While momentum and NAG improve the convergence rate and accuracy of SGD, they make it more difficult to scale to additional asynchronous workers. To simplify the analysis, we henceforth assume that all workers have equal computation power. This assumption can be relieved by monitoring the rate of each worker's updates and weighting them accordingly. We denote by $prev(i, t)$ the last iteration in which worker $i$ sent a gradient to the master before time $t$. For ASGD and NAG-ASGD\footnote{NAG-ASGD is \Cref{alg:asgd_master}, where the optimizer uses the aforementioned NAG method (\Cref{sec:background})}, $\E \left [\Delta_{t+\tau} \right ]$ is the sum of the gradients and the sum of the momentum vectors, respectively:
\begin{align}
    \label{eq:asgd_gap} \E[\Delta^\text{ASGD}_{t+\tau}] &= -\eta \sum_{i=1}^{N}g^{i}_{prev(i,t+\tau)} \\
    \label{eq:nag_asgd_gap} \E[\Delta^\text{NAG-ASGD}_{t+\tau}] &= -\eta \sum_{i=1}^{N}v_{prev(i,t+\tau)}
\end{align}
\Cref{fig:rmse_algos} demonstrates empirically that the \emph{gap} of NAG-ASGD is considerably larger than that of ASGD due to momentum\footnote{This is quite intuitive since generally the momentum vector is larger than the gradient.}, even though the \emph{lag} $\tau$ in both algorithms is exactly the same.

\subsection{Parameter Prediction} \label{sec:lwp}
The \textit{gap} arises from the fact that we can not access the future parameters $\theta_{t+\tau}$. \citet{kosson2020pipelined} proposed \textit{Linear Weight Prediction} (LWP) to estimate future parameters:
\begin{equation}
    \theta_{t+\tau} \approx \hat{\theta}^\text{LWP}_t \triangleq \theta^0_t-\tau\eta v_{t-1}
\end{equation}
To reduce the \emph{gap}, LWP approximate the master's future parameters $\theta_{t+\tau}$ in $\tau$ update steps. In other words, LWP scales the NAG estimation according to the number of times $v_{t-1}$ is used throughout the next $\tau$ updates. The worker then computes its gradient on $\hat{\theta}^\text{LWP}_t$ instead of $\theta^0_t$. LWP is frequently used in asynchronous model parallel pipelines \citep{guan2019xpipe,chen2018efficient,narayanan2019pipedream} to reduce the staleness in the backpropagation; during the forward pass LWP estimates the parameters of the backwards pass. \Cref{alg:LWP_master} describes the LWP master algorithm.
\begin{algorithm}[H]
\caption{LWP: master}
\label{alg:LWP_master}
\begin{algorithmic}
    \setlength{\itemindent}{-0.5em} 
    \STATE Receive gradient $g^i$ from worker $i$
    \STATE Update momentum $v \gets \gamma v+g^i$
    \STATE Update master's weights $\theta^0 \gets \theta^0-\eta v$
    \STATE Send estimate $\hat{\theta} = \theta^0-\tau\eta v$ to worker $i$
\end{algorithmic}
\end{algorithm}
However, in large-scale asynchronous settings LWP is not beneficial because as $\tau$ increases, the effect of $v_{t-1}$ on reaching $\theta_{t+\tau}$ diminishes. \Cref{fig:rmse_algos} shows that despite LWP parameter estimation $\hat{\theta}^\text{LWP}_t$, its \emph{gap} is still large and only slightly lower than NAG-ASGD.

DANA-Zero, detailed in the next section, maintains a small \emph{gap} throughout training, despite using momentum. The small \emph{gap} enables DANA-Zero to compute more accurate gradients, as given by \Cref{eq:lipschitz_gap}, and therefore achieve fast convergence and high final accuracy.

\section{DANA}
\label{sec:dana-section}
DANA (Distributed Adaptive NAG ASGD) is a distributed asynchronous technique that achieves state-of-the-art accuracy and fast convergence, even when trained with momentum on large clusters. DANA is designed to reduce the \emph{gap} by computing the gradient $g_t$ on parameters that approximate the master's future position $\theta_{t+\tau}$ using a similar \emph{look-ahead} to that of NAG. Thus, for the same \emph{lag}, DANA benefits from a reduced \emph{gap}, as shown by \Cref{fig:rmse_algos}, and therefore suffers less from gradient staleness.

\subsection{The DANA-Zero Update Rule} \label{sec:dana-zero}
To rectify the problem of LWP we propose to maintain at the master a separate momentum vector $v^i$ for each worker $i$; these vectors are updated exclusively with the corresponding worker's gradients $g^i$ using the same update rule as in vanilla SGD with momentum (\Cref{eq:momentum}). We refer to this method as \emph{Multi-ASGD} (\Cref{sec:algorithms}). 

To complete our adaptation of NAG to the distributed case, we propose to perform the look-ahead using the most recent momentum vectors of \emph{all} of the workers. We name this method \emph{DANA-Zero}. Instead of sending the master's current parameters $\theta^0_t$, DANA-Zero sends $\hat{\theta}^\text{DANA}_t$, the estimated future position of the master's parameters after the next $N$ updates, one for each worker:
\begin{align} 
     \label{eq:multi_momentum} v_{t}^i &\triangleq \gamma v_{prev(i,t-1)}^i+g_t^i \\
    \label{eq:dana_look_ahead} \hat{\theta}^\text{DANA}_t &\triangleq \theta^0_t-\eta\gamma \sum_{i=1}^{N}v^i_{prev(i, t-1)}
\end{align}
\begin{algorithm}[H]
\caption{DANA-Zero: master}
\label{alg:DANA-Zero_master}
\begin{algorithmic}
    \setlength{\itemindent}{-0.5em} 
    \STATE Receive gradient $g^i$ from worker $i$
    \STATE Update worker's momentum $v^i \gets \gamma v^i+g^i$
    \STATE Update master's weights $\theta^0 \gets \theta^0-\eta v^i$
    \STATE Send estimate $\hat{\theta} = \theta^0-\eta\gamma \sum_{j=1}^{N}v^j$ to worker $i$
\end{algorithmic}
\end{algorithm}
\Cref{alg:DANA-Zero_master} shows the DANA-Zero master algorithm. Unlike ASGD (\Cref{alg:asgd_master}), the master now sends back to the worker a future prediction of the parameters $\hat{\theta} = \theta^0-\eta\gamma \sum_{j=1}^{N}v^j$ instead of the current parameters $\theta^0$. This future prediction is what allows DANA-Zero to decrease the \emph{gap} and therefore compute accurate gradients. The worker remains the same as in ASGD (\Cref{alg:asgd_worker}). \Cref{fig:rmse_algos} shows that DANA-Zero accurately estimates the future parameters $\theta_{t+\tau}$ and therefore decreases its \emph{gap}. 

\Cref{thm:dana_gap} proves that $\E \left [\Delta^\text{DANA}_{t+\tau} \right] =\E \left [\Delta^\text{ASGD}_{t+\tau} \right ]$ in an asynchronous environment of $N$ equal computational powered workers:
\begin{equation} \label{thm:dana_gap}
\begin{split}
    \E \left [\Delta^\text{DANA}_{t+\tau} \right] =& \E \left [\theta_{t+\tau} \right ] - \E \left [\hat{\theta}_t \right ] \\
    \underbrace{=}_{\text{\Cref{eq:dana_look_ahead}}}& \theta_{t} - \eta \sum_{i=1}^{N}\left(v^i_{prev(i, t+\tau)}\right) \\
    &- \left (\theta_{t} - \eta\gamma \sum_{i=1}^{N}v^i_{prev(i,t-1)}\right )\\
    \underbrace{=}_{\text{\Cref{eq:multi_momentum}}}&\theta_{t} - \eta \sum_{i=1}^{N}\left(\gamma v^i_{prev(i,t-1)} + g^i_{prev(i,t+\tau)} \right)\\ 
    &- \left (\theta_{t} - \eta\gamma \sum_{i=1}^{N}v^i_{prev(i,t-1)}\right )\\
    =& -\eta \sum_{i=1}^{N}g^{i}_{prev(i,t+\tau)} \underbrace{=}_{\text{\Cref{eq:asgd_gap}}} \E \left [\Delta^\text{ASGD}_{t+\tau} \right ] \\
\end{split}
\end{equation}
\Cref{thm:dana_gap} shows that despite using momentum, DANA-Zero has a similar \emph{gap} to that of ASGD. \Cref{fig:rmse_algos} demonstrates this empirically: DANA-Zero maintains a small \emph{gap} throughout the training process. We note that, due to momentum, DANA-Zero converges faster than ASGD, resulting in smaller gradients, and therefore a smaller \emph{gap}. For more details and an empirical validation see \Cref{sec:normalized_gap}.

We note that computing the full summation $\sum_{i=1}^{N}v^i_{prev(i, t-1)}$ in DANA-Zero can be done in $\mathcal{O}(k)$, instead of $\mathcal{O}(k \cdot N)$ by maintaining $v^0 = \sum_{i=1}^{N}v^i_{prev(i, t-1)}$ and updating it using $v^0_t = v^0_t - v^i_{prev(i, t-1)} + v^i_{t}$. For further details please see \Cref{sec:efficient_dana}.

\paragraph{DANA-Zero Equivalence to Nesterov}
When running with one worker $(N=1)$ DANA-Zero reduces to a single NAG optimizer. This can be shown by merging the worker and master (\Cref{alg:asgd_worker,alg:DANA-Zero_master}) into a single algorithm: since at all times $\theta^1_t=\theta^0_t-\eta\gamma v_{t-1}$, the resulting algorithm trains one set of parameters $\theta$, which is exactly the NAG update rule. \Cref{alg:DANA-Zero_combined} shows the combined algorithm, equivalent to the standard NAG optimizer.
\begin{algorithm}[H]
	\caption{Fused DANA-Zero (when $N=1$)}
	\label{alg:DANA-Zero_combined}
	\begin{algorithmic}
		\STATE Compute gradient $g_t \gets \nabla J(\theta_t-\eta \gamma v_{t-1})$
		\STATE Update momentum $v_{t} \gets \gamma v_{t-1}+g_t$
		\STATE Update weights $\theta_{t+1} \gets \theta_t-\eta v_{t}$
	\end{algorithmic}
\end{algorithm}
\subsection{Optimizing DANA}
\label{sec:dana}
In DANA-Zero, the master maintains a momentum vector for every worker, and must also compute $\hat{\theta}^\text{DANA}$ at each iteration. This adds a small computation and memory overhead to the master. DANA-Slim, a variation of DANA-Zero, obtains the same look-ahead as DANA-Zero but without any computation or memory overhead. Thus, DANA-Slim maintains the same gradient staleness mitigation as DANA-Zero.

\paragraph{Bengio-NAG}
\citet{bengio} proposed a variation of NAG that simplifies the implementation and reduces computation cost. Known as Bengio-NAG, it defines a new variable $\Theta$ to stand for $\theta$ after the momentum update:
\begin{equation} \label{eq:bengio}
    \Theta_t \triangleq \theta_t -\eta\gamma v_{t-1} \\
\end{equation}
Substituting $\theta_t$ with $\Theta_t$ in the NAG update rule, using \Cref{eq:bengio}, yields the Bengio-NAG update rule:
\begin{equation}
\label{eq:bengio_2}
\begin{split}
    \theta_{t+1}&=\theta_t-\eta v_{t} \\
    \Theta_{t+1}+\eta \gamma v_{t}&=\Theta_t+\eta \gamma v_{t-1}-\eta v_{t} \\
    \Theta_{t+1}&= \Theta_t-\eta(\gamma v_{t}+\nabla J(\Theta_t; \xi)) 
\end{split}
\end{equation}
\Cref{eq:bengio_2} shows the Bengio-NAG update rule, where the gradient is both computed on and applied to $\Theta$, rather than computed on $\hat{\theta}$ but applied to $\theta$. Hence, an implementation of Bengio-NAG needs to store only one set of parameters $\Theta$ in memory and doesn't require computing $\hat{\theta}$.

\paragraph{The DANA-Slim Update Rule}
In creating DANA-Slim, we optimized DANA-Zero by leveraging the Bengio-NAG approach. We re-define $\Theta_{t}$ as $\theta_t$ after applying the momentum update from all future workers. Therefore, $\Theta_{t+1}$ is $\Theta_{t}$ after the current worker's update:
\begin{equation} \label{eq:dana_bengio_step}
\begin{split}
    \Theta_t &\triangleq \theta_t -\eta\gamma\sum_{j=1}^{N}v^j_{prev(j,t-1)} \\
    \Theta_{t+1} &= \theta_{t+1} -\eta\gamma\Big(v^i_{t}+\sum_{j \ne i}v^j_{prev(j, t-1)}\Big)
\end{split}
\end{equation}
DANA-Slim eliminates both the computational and memory overhead at the master by substituting $\theta_t$ with $\Theta_t$.
\begin{equation}
\label{eq:dana_slim_upd}
\begin{split}
    \theta_{t+1}&=\theta_t-\eta v^i_{t} \\
    &\Downarrow \text{\quad \Cref{eq:dana_bengio_step}}\\
    \Theta_{t+1} &+ \eta\gamma\Big(v^i_{t}+\sum_{j \ne i}^N v^j_{prev(j,t-1)}\Big)\\
    &=\Theta_t+\eta\gamma\sum_{j=1}^{N}v^j_{prev(j,t-1)}-\eta v^i_{t}\\
    \Theta_{t+1}&=\Theta_t+\eta\gamma\left (v^i_{prev(i,t-1)}-\left (1+\frac{1}{\gamma}\right )\cdot v^i_{t}\right ) \\
    &\Downarrow \text{\quad \Cref{eq:multi_momentum}}\\
    \Theta_{t+1}&=\Theta_t-\eta(\gamma v^i_{t}+\nabla J(\Theta_{prev(i,t)}; \xi)) 
\end{split}
\end{equation}
\Cref{eq:dana_slim_upd} shows that DANA-Slim benefits from the same gradient staleness mitigation as DANA-Zero up to a parameter switch. In DANA-Slim the master sends its current parameters $\Theta_t$ instead of computing the future parameters $\hat{\theta}$. Therefore, the master doesn't need to maintain the momentum vectors of all the workers.

\Cref{alg:DANA_worker} describes the worker algorithm of DANA-Slim. DANA-Slim only changes the worker algorithm, while using the same master algorithm as ASGD\footnote{Although \Cref{alg:asgd_master} receives $g_t$ while \Cref{alg:DANA_worker} sends $v_t$, there is no inconsistency since $v_t, g_t \in \mathbb{R}^k$.} (\Cref{alg:asgd_master}). Hence, it completely eliminates the overhead at the master and enjoys the same linear speedup scaling capabilities of ASGD. DANA-Slim is equivalent to DANA-Zero in all other aspects and provides the same gradient staleness mitigation.

\begin{algorithm}[H]
\caption{DANA-Slim: worker $i$}
\label{alg:DANA_worker}
\centering
\begin{algorithmic}
    \setlength{\itemindent}{-0.5em} 
	\STATE Receive parameters $\Theta^i$ from master 
	\STATE Compute gradient $g^i \gets \nabla J(\Theta^i; \xi)$
	\STATE Update momentum $v^i \gets \gamma v^i+g^i$
	\STATE Send update vector $\gamma v^i + g^i$ to master 
\end{algorithmic}
\end{algorithm}
\subsection{Delay Compensation} \label{sec:dc}
\citet{dcasgd} proposed \emph{Delay Compensated ASGD} (DC-ASGD), which tackles the problem of stale gradients by adjusting the gradient with a second-order Taylor expansion. Due to the high computation and space complexity of the Hessian matrix, they propose a cheap yet effective Hessian approximator, which is solely based on previous gradients. We denote by $\odot$ a matrix element-wise multiplication. 
\begin{equation} \label{eq:dc_asgd}
\begin{split}
    g_t = \nabla J(\theta_t; \xi) \quad ; \quad
    \hat{g}_t &= g_t + \lambda g_t \odot g_t \odot (\theta_{t+\tau} - \theta_t) \\
    \theta_{t+\tau+1} &= \theta_{t+\tau}-\eta \hat{g}_t
\end{split}
\end{equation}
\Cref{eq:dc_asgd} describes DC-ASGD. The delay compensation term, $\lambda g_t \odot g_t \odot (\theta_{t+\tau} - \theta_t)$, adjusts the gradient $g_t$ as if it were computed on $\theta_{t+\tau}$ instead of $\theta_t$; thus, mitigating the gradient staleness. However, DC-ASGD adds a memory overhead to the master since it now stores the previously sent parameters for each worker.

A Taylor expansion is more accurate when the source $\theta_t$ is in close vicinity to the approximation point $\theta_{t+\tau}$ (a small \emph{gap}). Momentum increases the \emph{gap}, thus reducing the effectiveness of DC-ASGD. DANA-Zero ensures that the \emph{gap} is kept small throughout training, even when using momentum; this amplifies the effectiveness of the delay compensation. The combined method, referred to as \emph{DANA with Delay Compensation} (DANA-DC), is described in \Cref{alg:DANA-DC_master}.
\begin{algorithm}[H]
\caption{DANA-DC: master}
\label{alg:DANA-DC_master}
\begin{algorithmic}
    \setlength{\itemindent}{-0.5em} 
    \STATE Receive gradient $g^i$ from worker $i$
    \STATE Update the gradient according to the delay compensation term $\hat{g}^i = g^i + \lambda g^i \odot g^i \odot (\theta^0 - \theta^i)$
    \STATE Update momentum $v^i \gets \gamma v^i+\hat{g}^i$
    \STATE Update master's weights $\theta^0 \gets \theta^0-\eta v^i$
    \STATE Send estimate $\hat{\theta} = \theta^0-\eta\gamma \sum_{j=1}^{N}v^j$ to worker $i$
\end{algorithmic}
\end{algorithm}
\section{Experiments} \label{sec:experiments}
In this section, we present our evaluations and insights regarding DANA. In \Cref{sec:cifar,sec:imagenet} we focus on accuracy rather than communication overheads, and therefore we simulate multiple distributed workers\footnote{A single worker may be more than a single GPU. DANA, like all ASGD algorithms, treats each machine with multiple GPUs as a single worker. For example, DANA can run on 32 workers with 8 GPUs each (256 GPUs in total), where each worker performs SSGD internally, which is transparent to the ASGD algorithm.} and measure the final test error and convergence speed of different cluster sizes. In \Cref{sec:gap_importance} we show the importance of decreasing the \emph{gap} in asynchronous environments and point to the high correlation between small \emph{gap} and high final test accuracy. Finally, in \Cref{sec:speedup_private,sec:speedup_public} we present real-world distributed asynchronous results that are trained in two settings: dedicated private compute cluster and public cloud data-center (Google cloud). We show that DANA trains over $25\%$ faster than an optimized SSGD algorithm while maintaining high final test accuracy and fast convergence.
\begin{figure*}[t]
\centering
    \begin{subfigure}{0.49\textwidth}
            \includegraphics[width=\columnwidth]{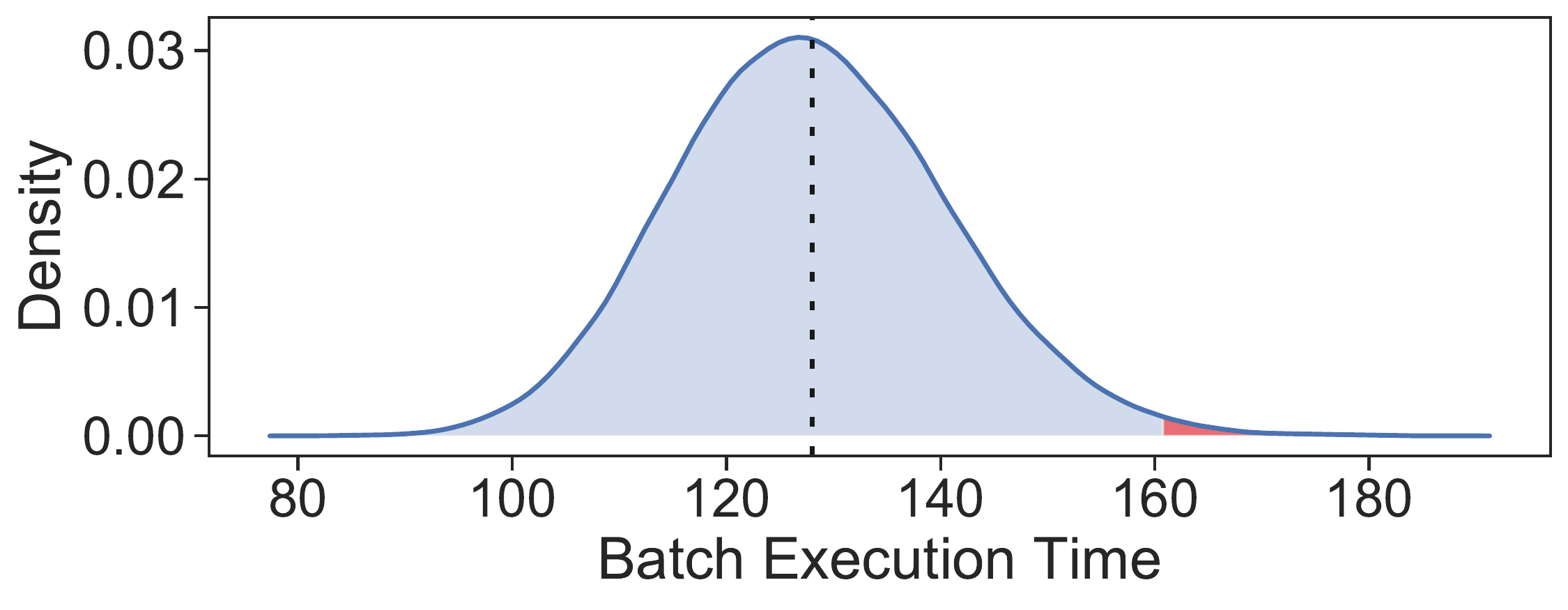}
        	\caption{Homogeneous \emph{gamma distribution} model.}
        	\label{fig:gamma_distribution_homo}
    \end{subfigure}
    \hfill
    \begin{subfigure}{0.49\textwidth}
            \includegraphics[width=\columnwidth]{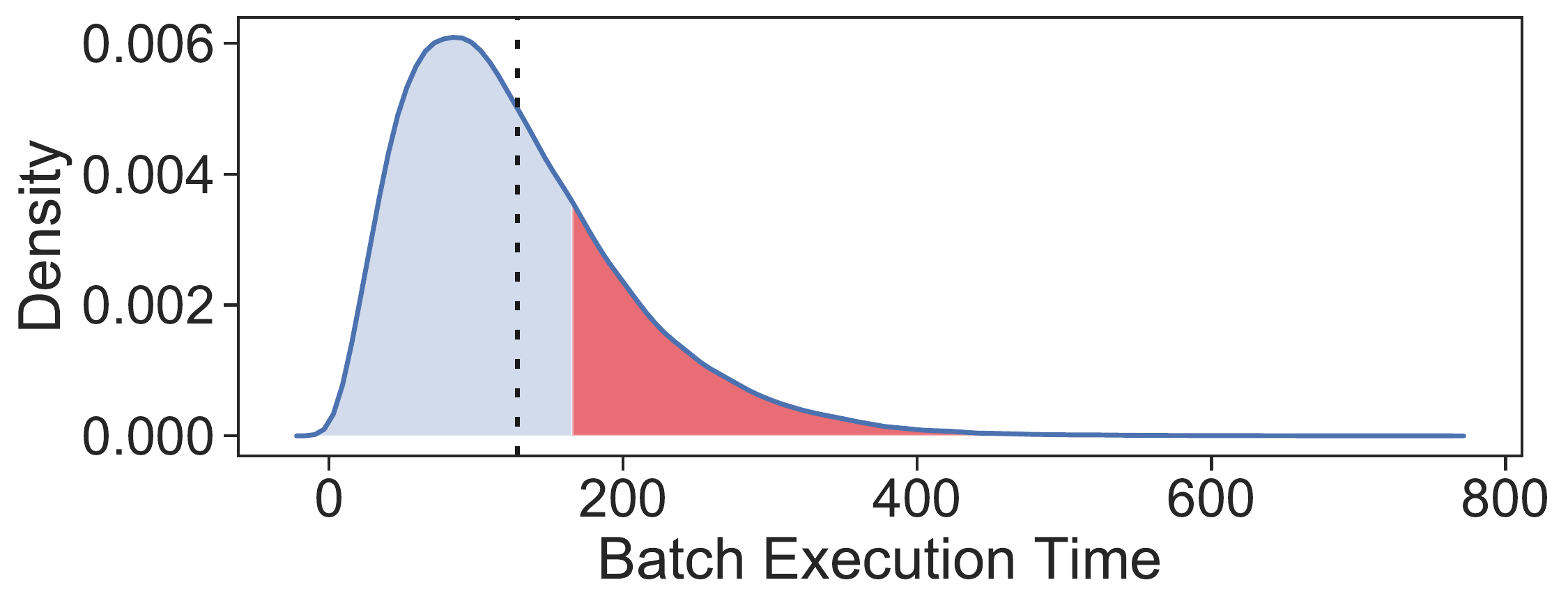}
        	\caption{Heterogeneous  \emph{gamma distribution} model.}
        	\label{fig:gamma_distribution_hetero}
    \end{subfigure}
    \caption{Gamma-distribution in homogeneous and heterogeneous environments. The \emph{x-axis} is the simulated time units the iteration takes while the \emph{y-axis} is the probability. Both environments have the same mean (128 time units). The red area represents the probability to have an iteration which takes more than 1.25x longer than the mean iteration time.}
\label{fig:gamma_distribution}
\end{figure*}
\begin{figure*}[t]
    \centering
    \begin{subfigure}{0.33\textwidth}
            \includegraphics[width=\textwidth]{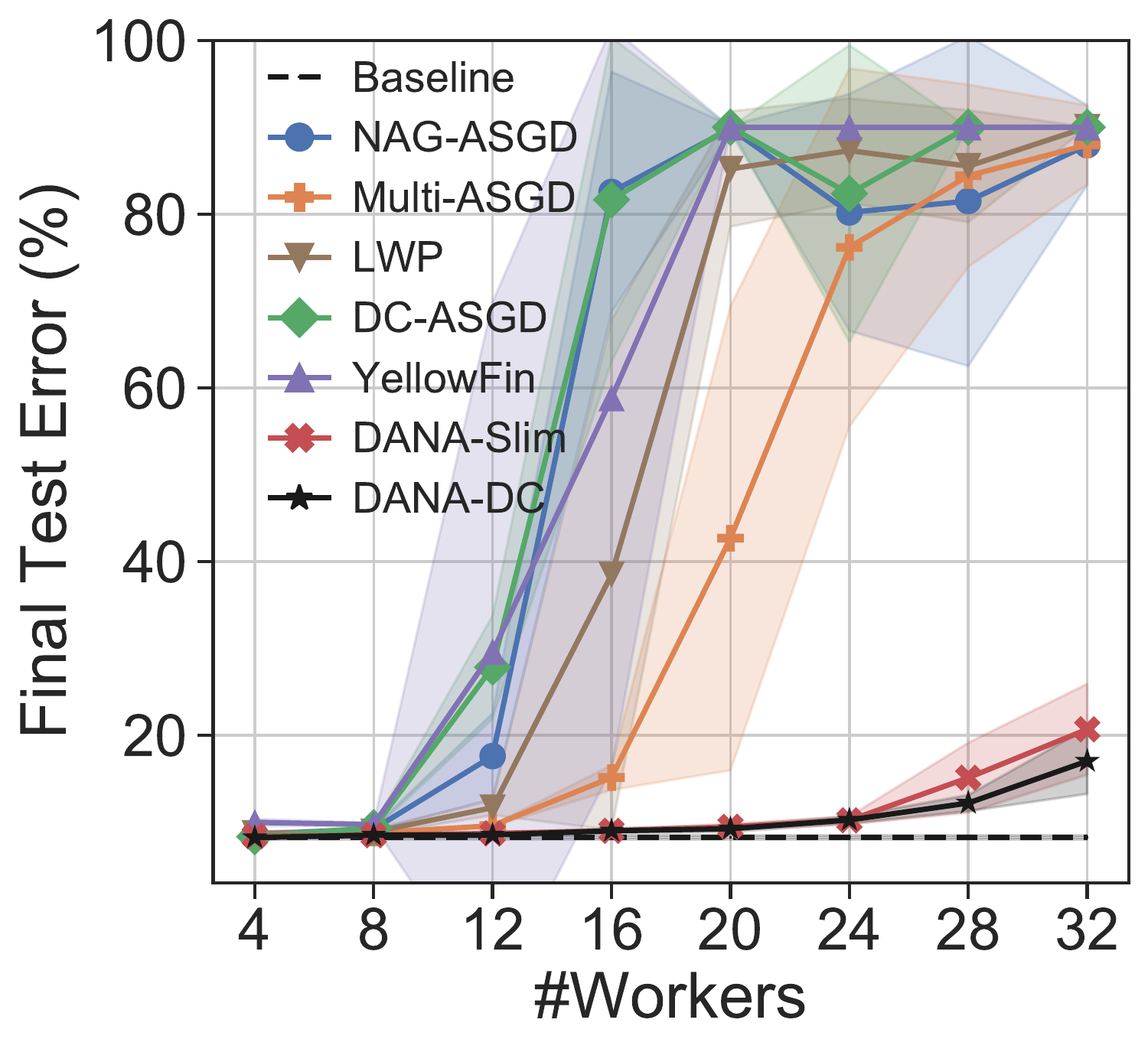}
            \caption{CIFAR10 ResNet-20}
        	\label{fig:acc_cifar10_resnet}
    \end{subfigure}
    \hfill
    \begin{subfigure}{0.33\textwidth}
            \includegraphics[width=\textwidth]{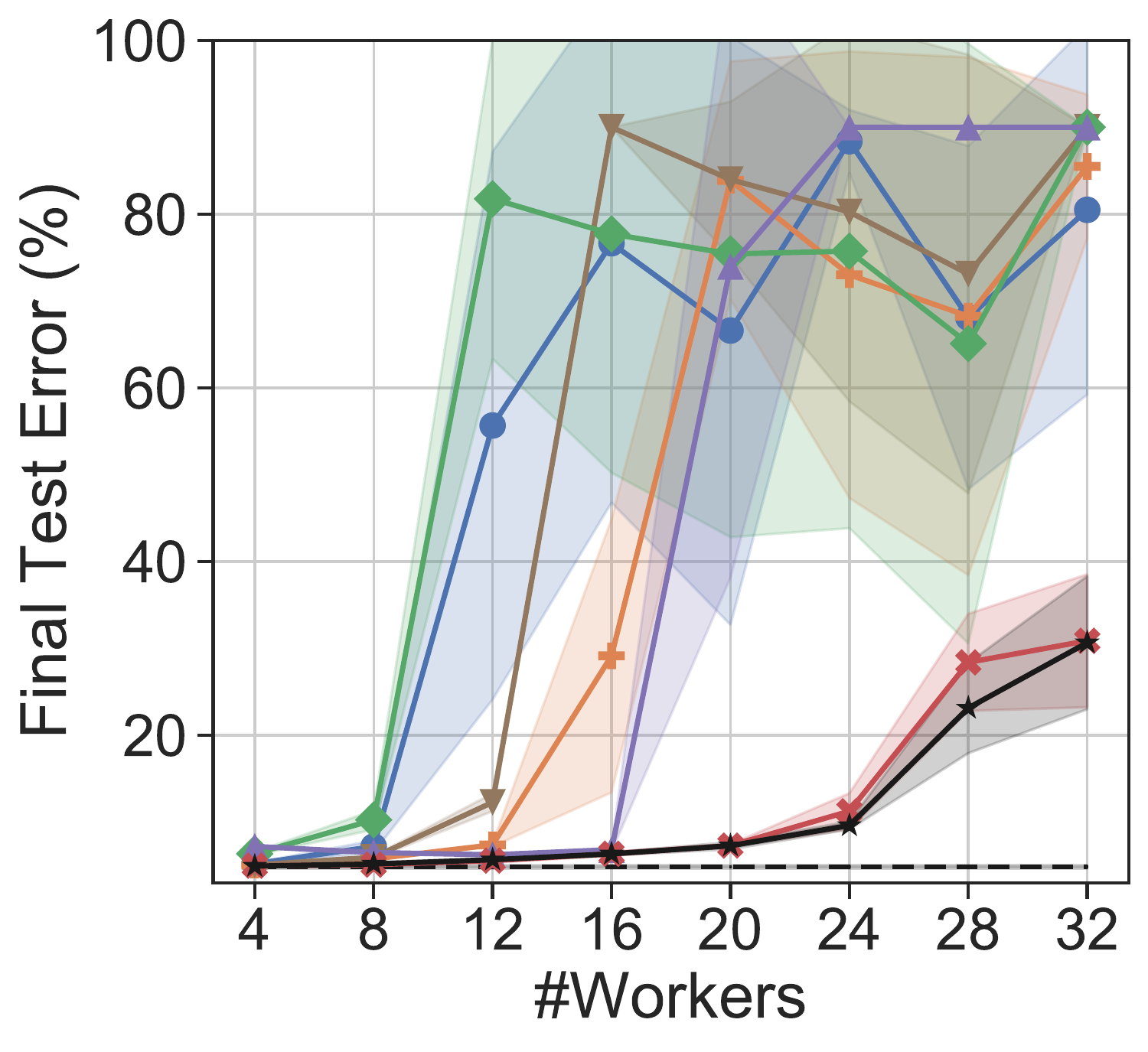}
        	\caption{CIFAR10 Wide ResNet 16-4}
        	\label{fig:acc_cifar10_wr}
    \end{subfigure}
    \hfill
    \begin{subfigure}{0.33\textwidth}
            \includegraphics[width=\textwidth]{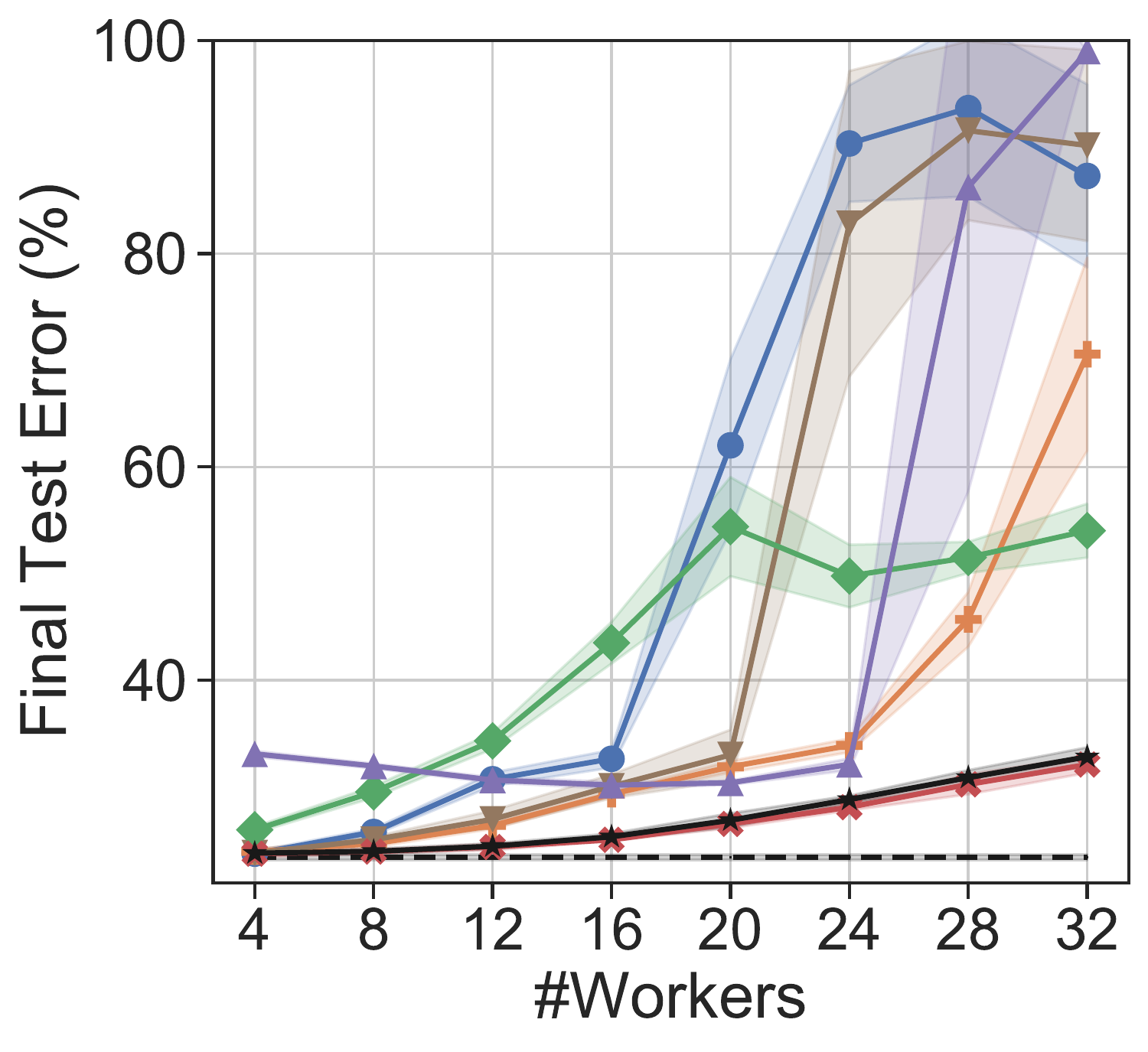}
            \caption{CIFAR100 Wide ResNet 16-4}
        	\label{fig:acc_cifar100_wr}
    \end{subfigure}
    \caption{Final test error for different numbers of workers $N$.}
    \label{fig:acc_cifar}
\end{figure*}
\paragraph{Simulation} We simulate the workers' execution time using a \emph{gamma-distributed model} \citep{Ali:2000:TET}, where the execution time for each individual batch is drawn from a gamma distribution. The gamma distribution is a well-accepted model for task execution time that naturally gives rise to stragglers. We use the formulation proposed by~\cite{Ali:2000:TET} and set $V=0.1$ and $\mu=B*V^2$, where $B$ is the batch size, yielding a mean execution time of $B$ simulated time units (additional details in \Cref{sec:gamma_distribution}). \Cref{fig:gamma_distribution} visualizes the distribution of workers' batch execution time. As expected, stragglers appear much more frequently in the heterogeneous environment than in the homogeneous environment.

Since one of our main goals in these experiments is to verify that decreasing the \emph{gap} leads to a better final test error and convergence rate, we use the same hyperparameters across all algorithms. These are the original hyperparameters suggested by the authors of each neural network architecture's respective paper, tuned for a single worker (for additional details please see \Cref{sec:hyperparameters}). This eases the scaling to more workers since it doesn't require re-tuning the hyperparameters when increasing the cluster size.

\paragraph{Algorithms} Our goal is to produce a scalable method that works well with momentum and therefore all evaluated algorithms use momentum (as does the baseline). Our evaluations consist of the following algorithms:
\begin{itemize}
	\item \emph{Baseline:} Single worker with the same tuned hyperparameters as in the respective neural network's paper. This baseline does not suffer from gradient staleness, thus it is ideal in terms of final accuracy and convergence speed.
	\item \emph{NAG-ASGD:} Asynchronous SGD, which uses a single NAG optimizer for all workers.
	\item \emph{Multi-ASGD:} Asynchronous SGD, which maintains a separate NAG optimizer for each worker.
	\item \emph{DC-ASGD:} Delay compensation asynchronous SGD, as described in \Cref{sec:dc}, for which we set $\gamma=0.95$ as suggested by \citet{dcasgd}.
	\item \emph{YellowFin:} An algorithm proposed by \citet{mlsys2019_153}, that automatically tunes the momentum $\gamma$ and learning rate $\eta$ throughout the training process. We used the asynchronous variation of YellowFin, named Closed-Loop, and set the hyperparameters $\eta = 1e-4$ and $\gamma = 0.0$ as suggested by \citet{mlsys2019_153}.
	\item \emph{LWP:} Linear Weight Prediction, described in \Cref{sec:lwp}.
	\item \emph{DANA-Slim:} A variation of DANA-Zero, which eliminates the overhead, as described in \Cref{sec:dana}.
	\item \emph{DANA-DC:} A combination of DANA-Zero with DC-ASGD, as described in \Cref{sec:dc}, for which we set $\lambda = 2$, as suggested by \citet{dcasgd}. 
\end{itemize}

\subsection{Evaluation on the CIFAR Datasets} \label{sec:cifar}
\begin{figure*}[t]
    \centering
    \begin{subfigure}{0.33\textwidth}
            \includegraphics[width=\textwidth]{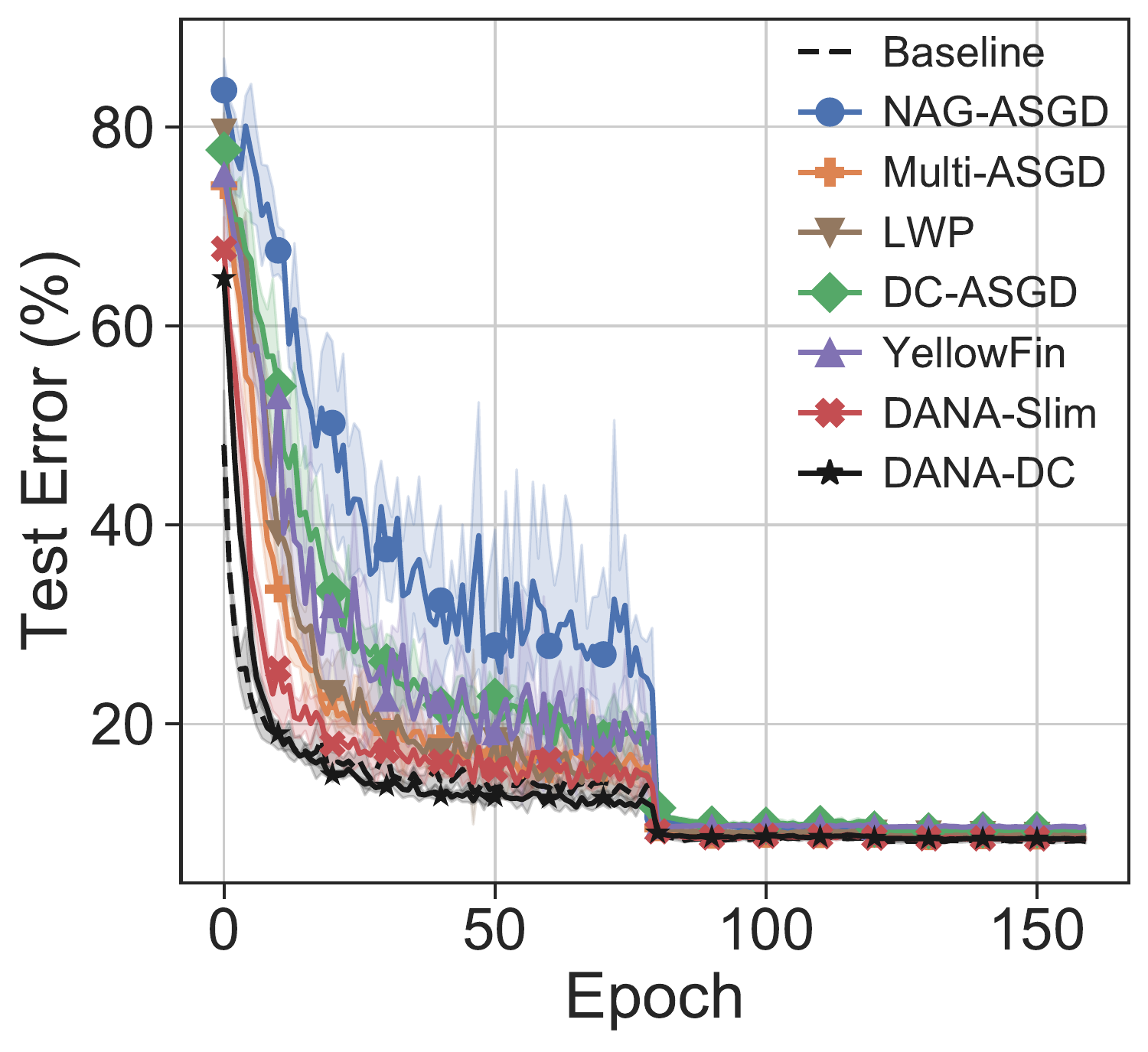}
            \caption{CIFAR10 ResNet-20}
        	\label{fig:conv_cifar10_resnet}
    \end{subfigure}
    \hfill
    \begin{subfigure}{0.33\textwidth}
            \includegraphics[width=\textwidth]{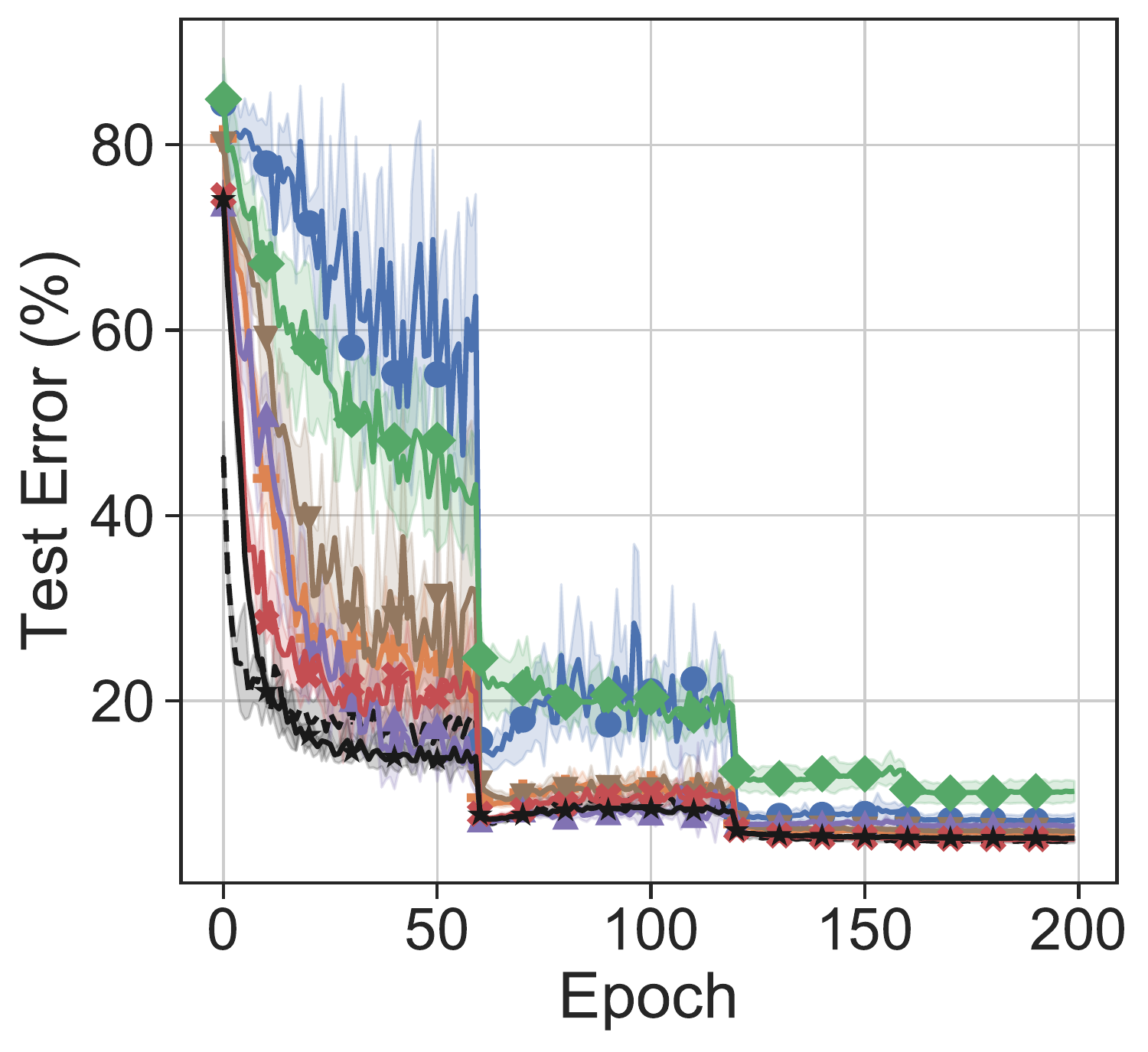}
        	\caption{CIFAR10 Wide ResNet 16-4}
        	\label{fig:conv_cifar10_wr}
    \end{subfigure}
    \hfill
    \begin{subfigure}{0.33\textwidth}
            \includegraphics[width=\textwidth]{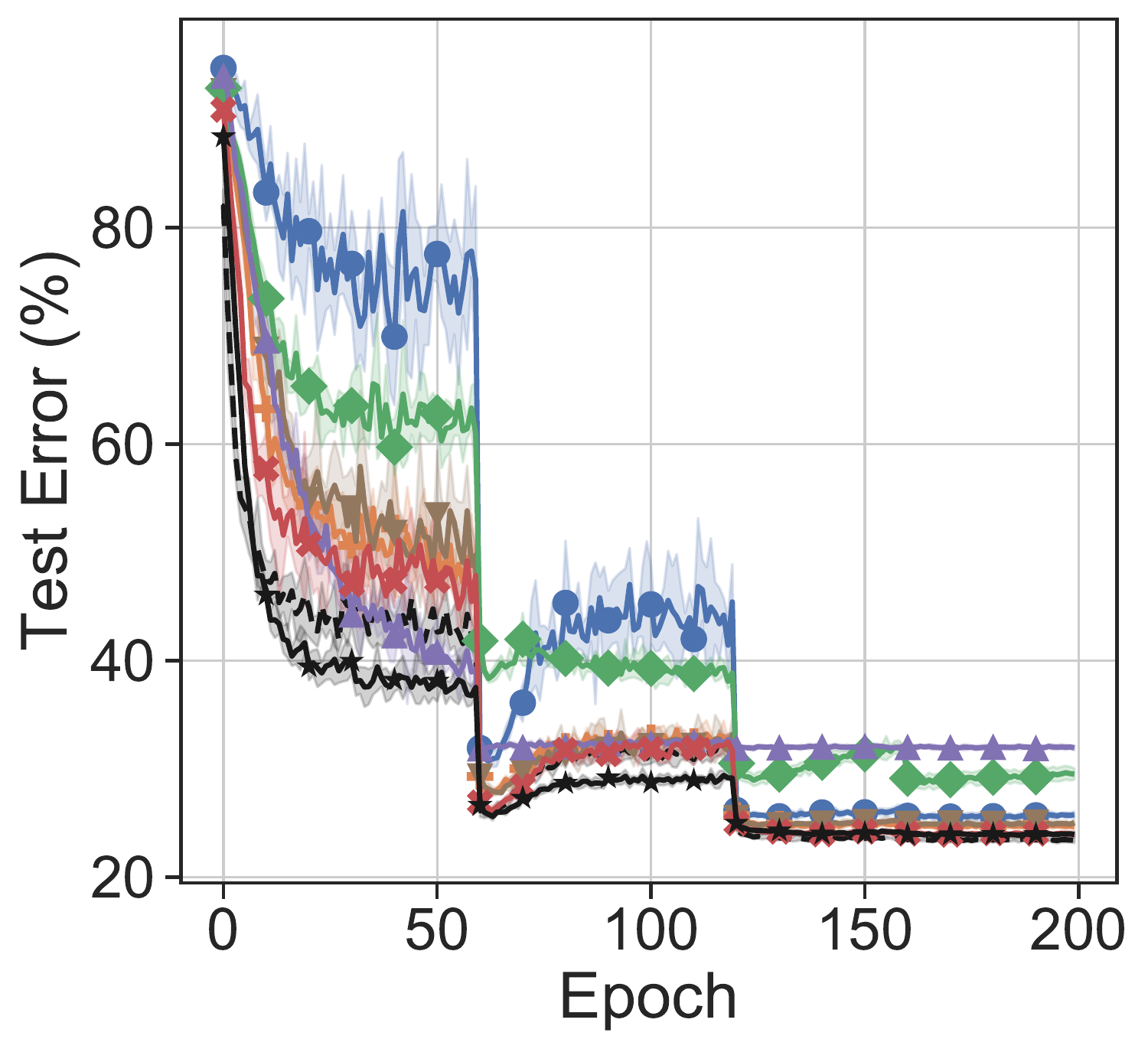}
            \caption{CIFAR100 Wide ResNet 16-4}
        	\label{fig:conv_cifar100_wr}
    \end{subfigure}
    \caption{Convergence rate for 8 workers.}
    \label{fig:conv_cifar}
\end{figure*}
We evaluate DANA with the ResNet-20 \citep{resnet} and Wide ResNet 16-4 \citep{wide_resnet} architectures on the CIFAR-10 and CIFAR-100 datasets \citep{cifar}. In the CIFAR experiments, bold lines show the mean over five different runs with random seeds, while transparent bands show the standard deviation. The baseline is the mean of five different runs with a single worker.

\Cref{fig:acc_cifar} shows that the final test error of both DANA-Slim and DANA-DC is lower than all the other algorithms for any number of workers, especially for large numbers of workers. Both variations of DANA exhibit a very small standard deviations, which points to the high stability DANA provides, even for large numbers of workers. The final accuracies of all the CIFAR experiments are listed in \Cref{sec:cifar_final}.

NAG-ASGD demonstrates how gradient staleness is exacerbated by momentum. NAG-ASGD yields good accuracy with few workers, but fails to converge when trained with more than 16 workers. LWP scales better than NAG-ASGD but falls short of Multi-ASGD. Multi-ASGD serves as an ablation study: its poor scalability demonstrates that it is not sufficient to simply maintain a momentum vector for every worker. Hence, DANA (\Cref{sec:dana-section}) is also required to achieve fast convergence and low test error.

\Cref{fig:conv_cifar} shows the mean and standard deviation of the test error throughout the training of the different algorithms when trained on eight workers. This figure demonstrates the significantly better convergence rate of DANA-DC. It is usually similar to the baseline or even faster and outperforms all the other algorithms. It is noteworthy that DANA-DC's convergence rate surpasses that of DANA-Slim; however, DANA-DC incurs an overhead and both algorithms usually reach a similar final test error, as seen in \Cref{fig:acc_cifar}.

\begin{figure}[h]
    \includegraphics[width=\columnwidth]{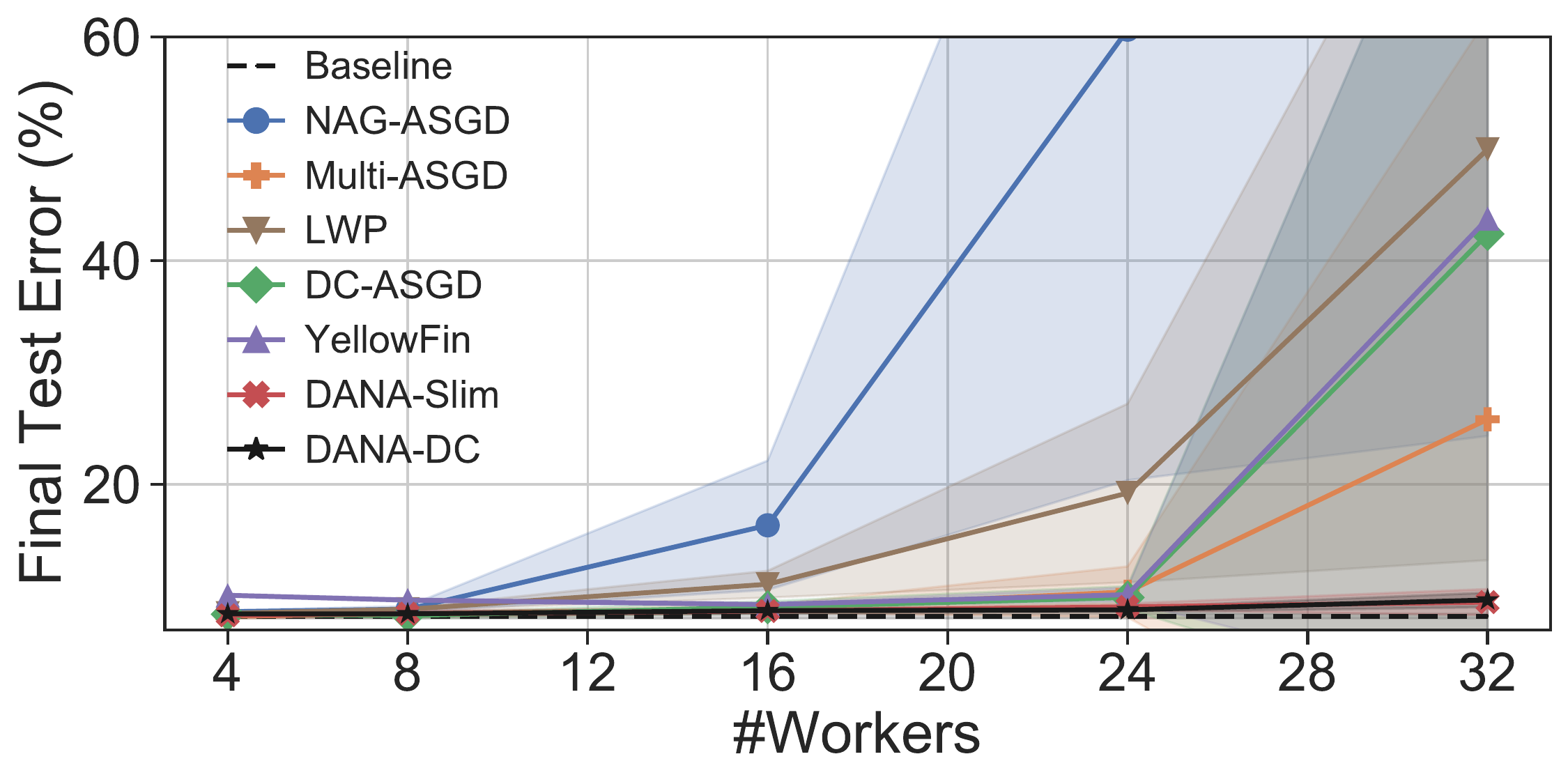}
	\caption{Final test error for different numbers of workers $N$ when training ResNet-20 on CIFAR10 in a heterogeneous environment.}
	\label{fig:hetero_acc_main}
\end{figure}
When attempting to utilize different computational resources \citep{yang2018adaptive,yang2020boa,woodworth2020minibatch}, workers may have considerably different computational power from one another. This heterogeneous environment creates high variance in batch execution times, as shown in \Cref{fig:gamma_distribution_hetero}. Therefore, in heterogeneous environments, asynchronous algorithms have a distinct speedup advantage over synchronous algorithms which are slowed down by the slowest worker if not addressed \citep{dutta18a,hanna2020adaptive}. \Cref{fig:hetero_acc_main} shows that even in heterogeneous environments, DANA achieves high final accuracy on large clusters of workers. For additional details about the heterogeneous experiments please see \Cref{sec:hetero}.

\subsection{Evaluation on the ImageNet Dataset}
\label{sec:imagenet}
\begin{figure*}[t]
    \centering
    \begin{subfigure}{0.49\textwidth}
            \includegraphics[width=\textwidth]{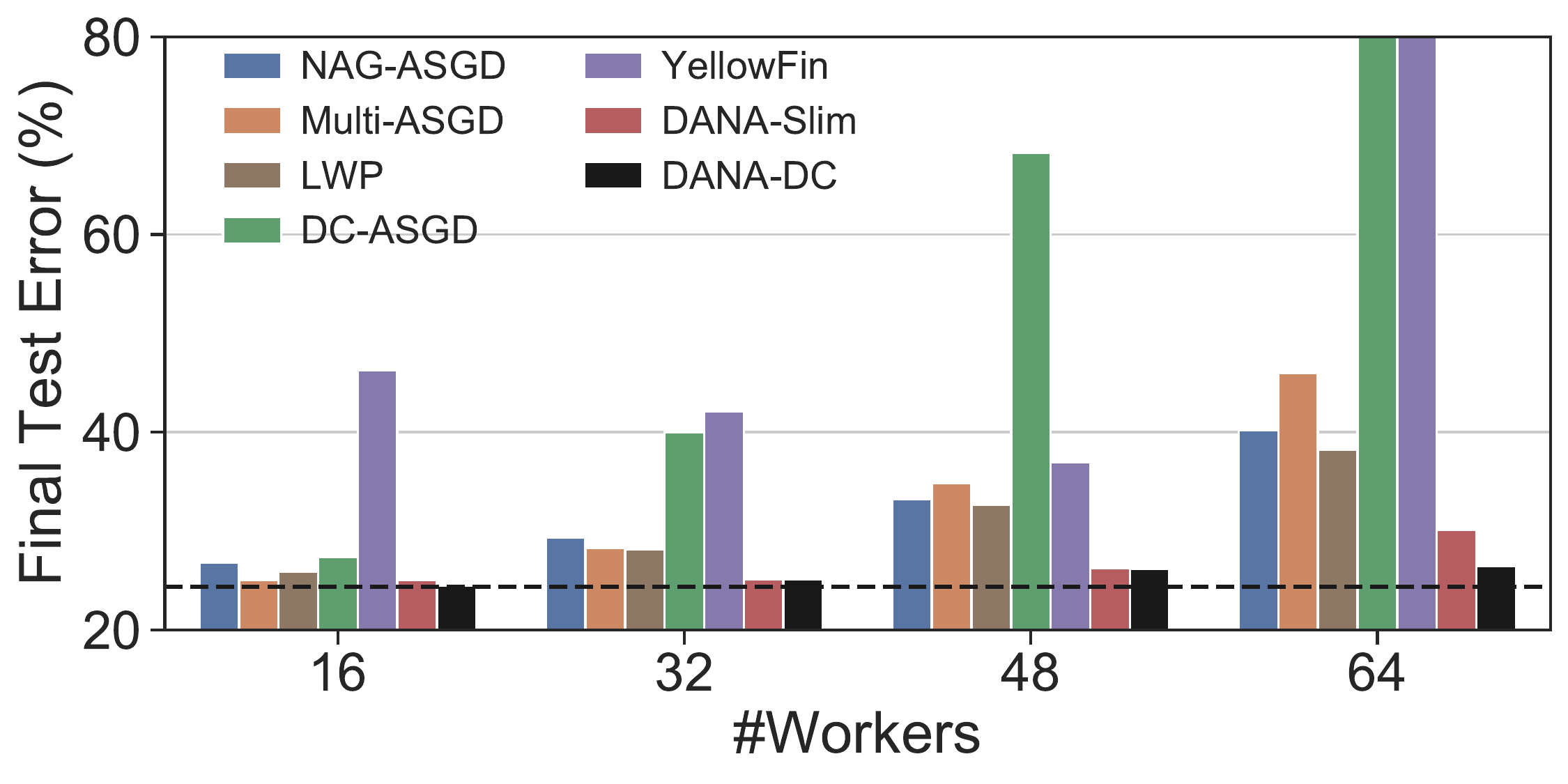}
            \caption{Final test error on 16, 32, 48, and 64 workers}
            \label{fig:imagenet_acc}
    \end{subfigure}
    \hfill
    \begin{subfigure}{0.49\textwidth}
        	\includegraphics[width=\textwidth]{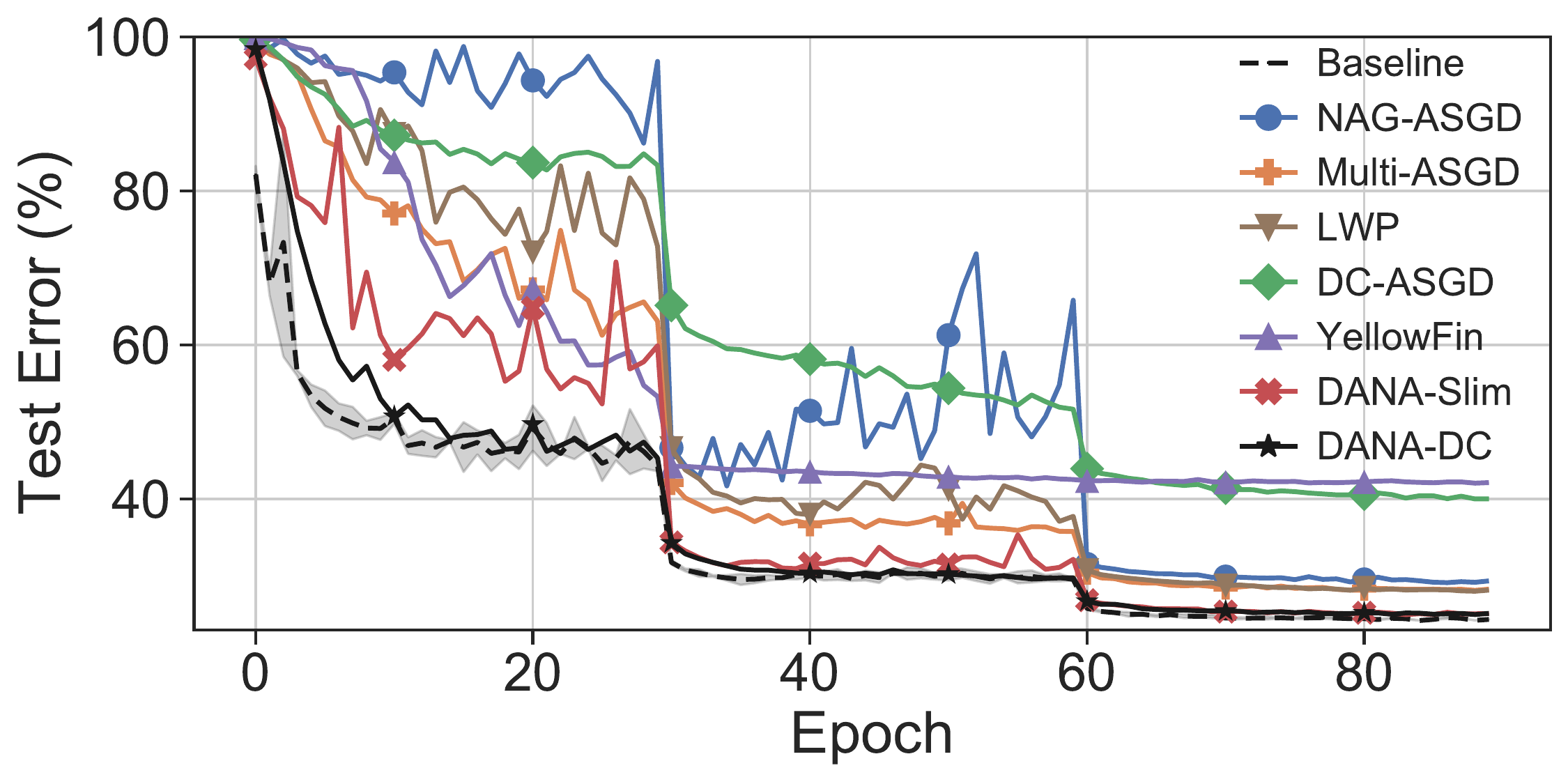}
        	\caption{Convergence rate on 32 workers}
        	\label{fig:imagenet_conv}
    \end{subfigure}
    \caption{\subref{fig:imagenet_acc}~and~\subref{fig:imagenet_conv} show the final test errors and convergence speed, respectively, when training a ResNet-50 architecture on the ImageNet dataset. The black dashed line represents the baseline of a single worker.}
    \label{fig:imagenet}
\end{figure*}
\Cref{fig:imagenet} shows experiments where we evaluate DANA with ResNet-50 \citep{resnet} on the ImageNet dataset \citep{imagenet}. In this experiment, every asynchronous worker is a machine with 8 GPUs, so the 64-worker experiment simulates a total of 512 GPUs with a total batch size of 16K.
\Cref{fig:imagenet_acc} compares final test errors on different cluster sizes and shows that the scalability in terms of final accuracy of DANA-Slim and DANA-DC is much greater than all the other algorithms. \Cref{fig:imagenet_conv} shows that both variations of DANA significantly outperform all the other algorithms in both convergence speed and final accuracy when trained on $32$ asynchronous workers. \Cref{sec:imagenet_final_accuracy} lists the final test accuracies on ImageNet.

\subsection{The Importance of the Gap} \label{sec:gap_importance}
Although all algorithms share the same average \emph{lag} (for a given number of workers), the algorithms that achieve a lower average \emph{gap} (\Cref{fig:rmse_algos}) also demonstrate low final error (\Cref{fig:acc_cifar}) and fast convergence rate (\Cref{fig:conv_cifar}). Ergo, we conclude that the \emph{gap} is more informative than the \emph{lag} when battling gradient staleness and that \emph{gap} reduction is paramount to asynchronous training. We note that the average \emph{gap} of both DANA-Zero and DANA-DC is an order of magnitude smaller than that of NAG-ASGD and LWP, as shown in \Cref{fig:rmse_algos}. Further details of our analysis on the \emph{gap} are discussed in \Cref{sec:normalized_gap}.

\subsection{Evaluation on a Private Compute Cluster} \label{sec:speedup_private}
While this work focuses on improving the accuracy of ASGD, we also measured acceleration in training time. We conduct our experiments on a system with $8$ Nvidia 2080ti GPUs that each have 11GB and base our code on PyTorch~\cite{paszke2019pytorch}. We use an efficient implementation for synchronous training\footnote{\url{https://github.com/pytorch/examples/tree/master/imagenet}} (\emph{DistibutedDataParallel} known as SSGD) based on the Ring-AllReduce communication algorithm that is implemented by Nvidia's NCCL collective communication package. Furthermore, in SSGD, we overlap its computations with communications to accelerate the training by over $15\%$ \citep{li2020pytorch}.

\begin{figure}[h]
    \centering
    \includegraphics[width=\columnwidth]{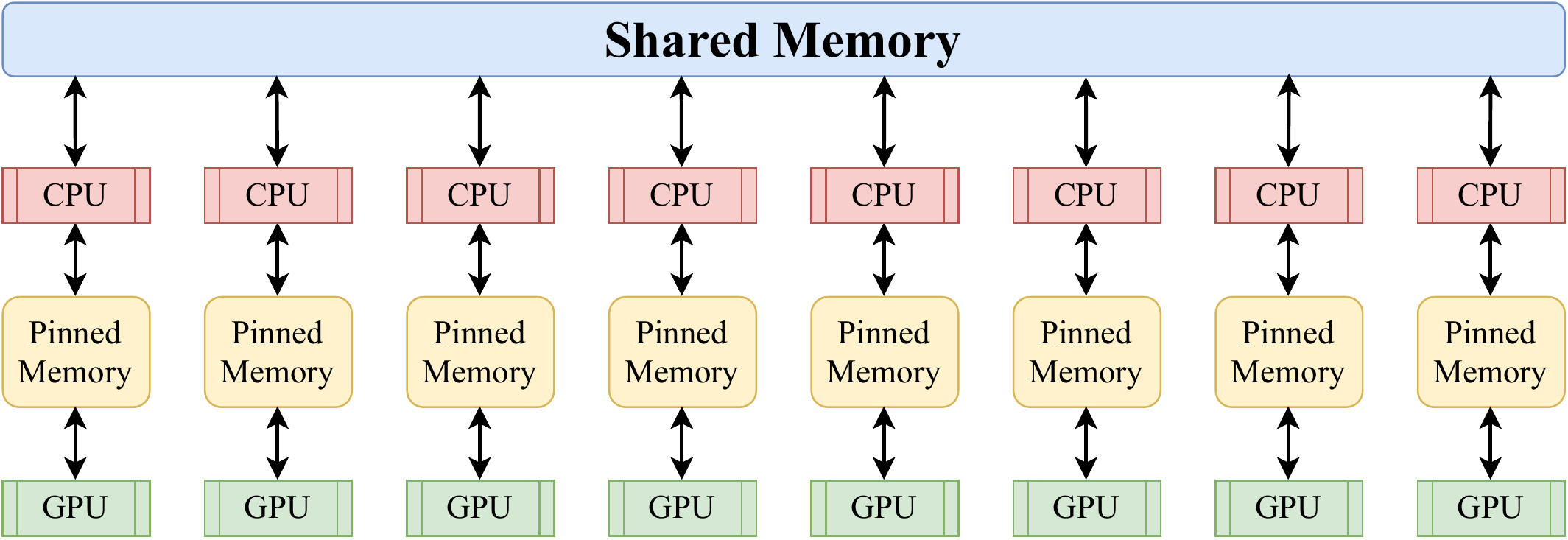}
	\caption{Our asynchronous lock-free training setup.}
	\label{fig:shared_memory}
\end{figure}
\begin{figure*}[t]
    \centering
    \begin{subfigure}{0.49\textwidth}
            \includegraphics[width=\textwidth]{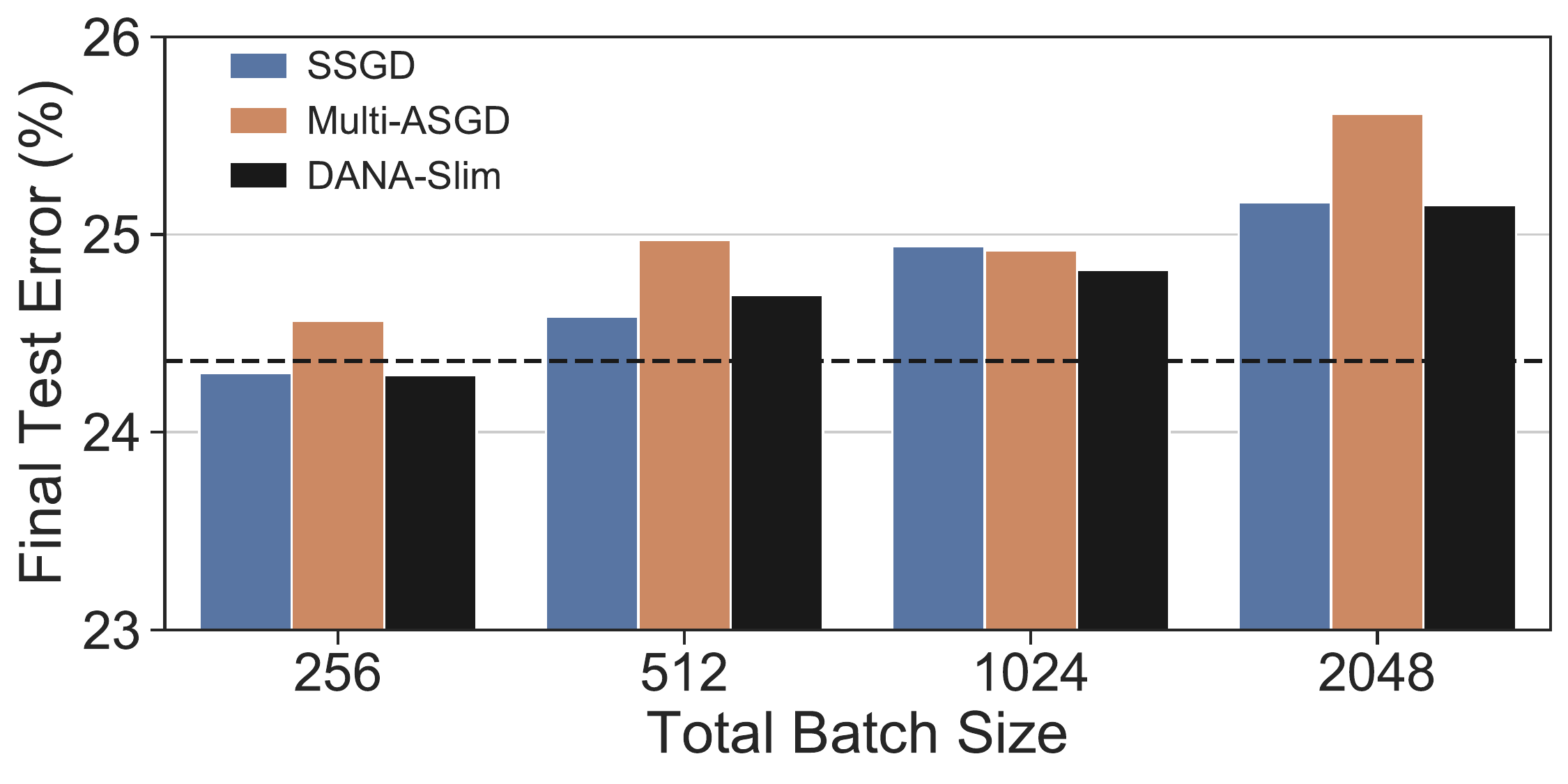}
            \caption{Final test error on 256, 512, 1024, and 2048 total batch sizes}
            \label{fig:distributed_imagenet_acc}
    \end{subfigure}
    \hfill
    \begin{subfigure}{0.49\textwidth}
        	\includegraphics[width=\textwidth]{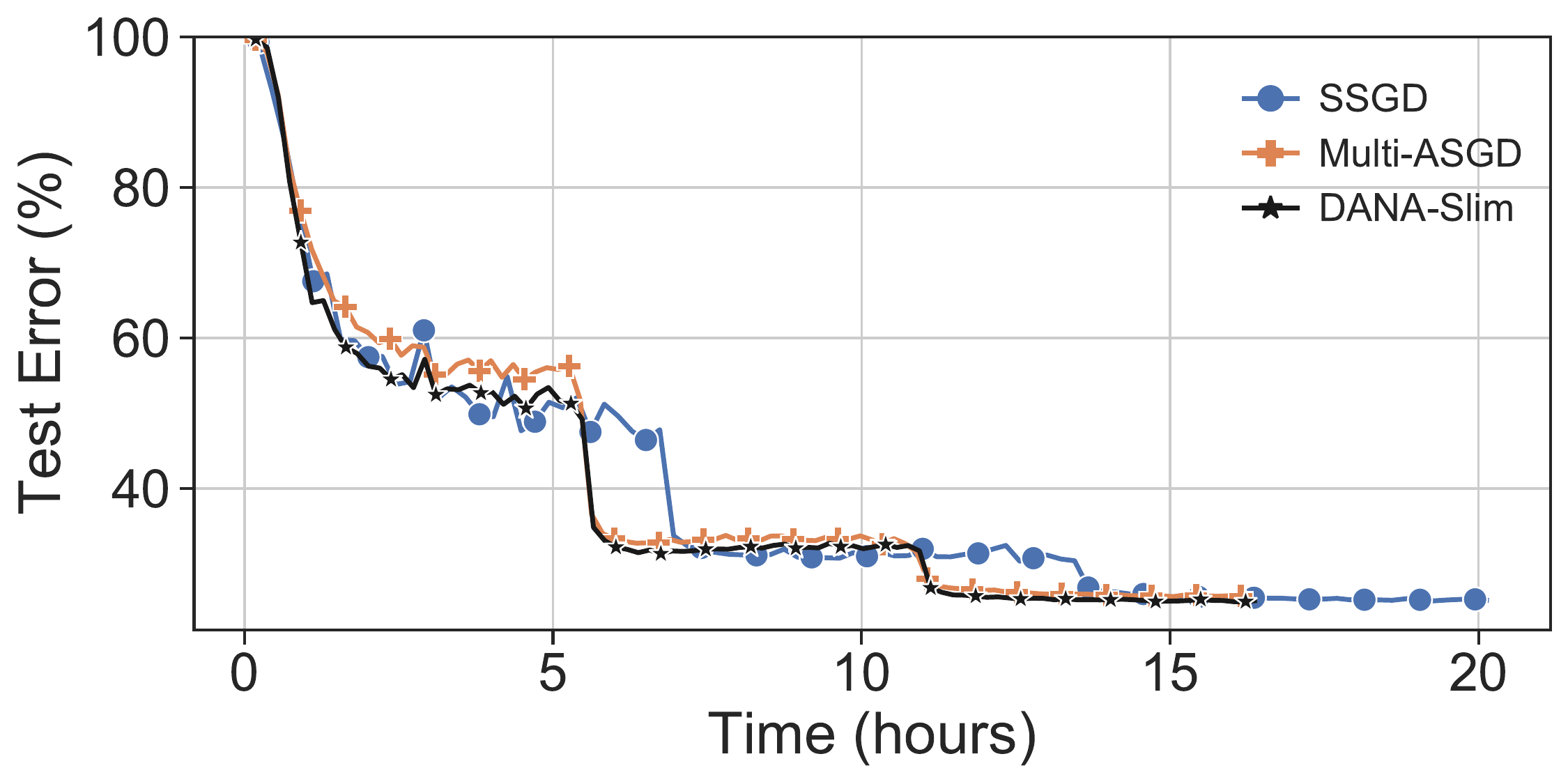}
        	\caption{Convergence rate on a total batch size of 2048}
        	\label{fig:distributed_imagenet_conv}
    \end{subfigure}
    \caption{ImageNet ResNet-50 distributed training with 8 GPUs. The black dashed line in \subref{fig:distributed_imagenet_acc} represents the baseline with a single worker.}
    \label{fig:disributed_speedup}
\end{figure*}
\begin{table*}[t]
\captionof{table}{Distributed results of ResNet-50 on ImageNet with an $8$ GPU machine (time in hours)}
\label{tab:distributed_imagenet_table}
\centering
\begin{tabular}{c c c c c c c c c c}
\toprule
& \multicolumn{3}{c}{\textbf{DANA-Slim}} & \multicolumn{3}{c}{\textbf{Multi-ASGD}} & \multicolumn{3}{c}{\textbf{SSGD}} \\
Total Batch Size & Accuracy & Time & Speedup & Accuracy & Time & Speedup & Accuracy & Time & Speedup\\
\midrule
256 
& \textbf{75.71\%} &  \textbf{20.3} & \textbf{6.78x}
& 75.44\% & 20.5 & 6.72x
& 75.7\% & 25.5 & 5.40x\\
\midrule
512
& 75.31\% &  \textbf{18} & \textbf{7.65x}
& 75.03\% & \textbf{18} & \textbf{7.65x}
& \textbf{75.42\%} & 22.9 & 6.01x\\
\midrule
1024 
& \textbf{75.18\%} &  \textbf{16.9} & \textbf{8.15x} 
& 75.08\% & \textbf{16.9} & \textbf{8.15x}
& 75.06\% & 20.9 & 6.59x\\
\midrule
2048 
& \textbf{74.85\%} & 16.4 & 8.39x
& 74.39\% & \textbf{16.3} &\textbf{ 8.45x}
& 74.84\% & 20.17 & 6.83x\\
\bottomrule
\end{tabular}
\end{table*}
\Cref{fig:shared_memory} shows our asynchronous training setup. Each worker is executed on a different process with a dedicated GPU. The shared parameters are stored and updated on the shared memory similar to Hogwild~\citep{recht2011hogwild}. The communications between the GPU and shared memory are transferred via pinned memory for fast parallel data transfer.

\Cref{fig:disributed_speedup} presents results when training the ResNet-50 architecture on ImageNet with different total batch sizes where every asynchronous worker is a single GPU. For total batch sizes that are larger than $256$ (batch size of $32$ per GPU) we use gradient accumulation \cite{aji-heafield-2019-making} to reduce memory footprint as well as reduce the synchronization frequency and communication overhead. Therefore, larger total batch sizes result in higher communication efficiency and shorter training times. \Cref{fig:distributed_imagenet_acc} compares the final test error on different total batch sizes. Multi-ASGD quickly drops in final accuracy when scaling the total batch size. DANA-Slim, on the other hand, not only scales well but in some cases even surpasses the final accuracy of SSGD. \Cref{fig:distributed_imagenet_conv} shows that DANA-Slim converges faster than both Multi-ASGD and SSGD. DANA-Slim trains $25\%$ faster than SSGD while achieving similar final accuracy. \Cref{tab:distributed_imagenet_table} lists the  final accuracies, training time, and the speedup over a single GPU. DANA-Slim achieves perfect linear scaling while maintaining final accuracy similar to the baseline.

\subsection{Evaluation on a Public Cloud Data-center} \label{sec:speedup_public}
\begin{figure}[h]
    \centering
    \includegraphics[width=\columnwidth]{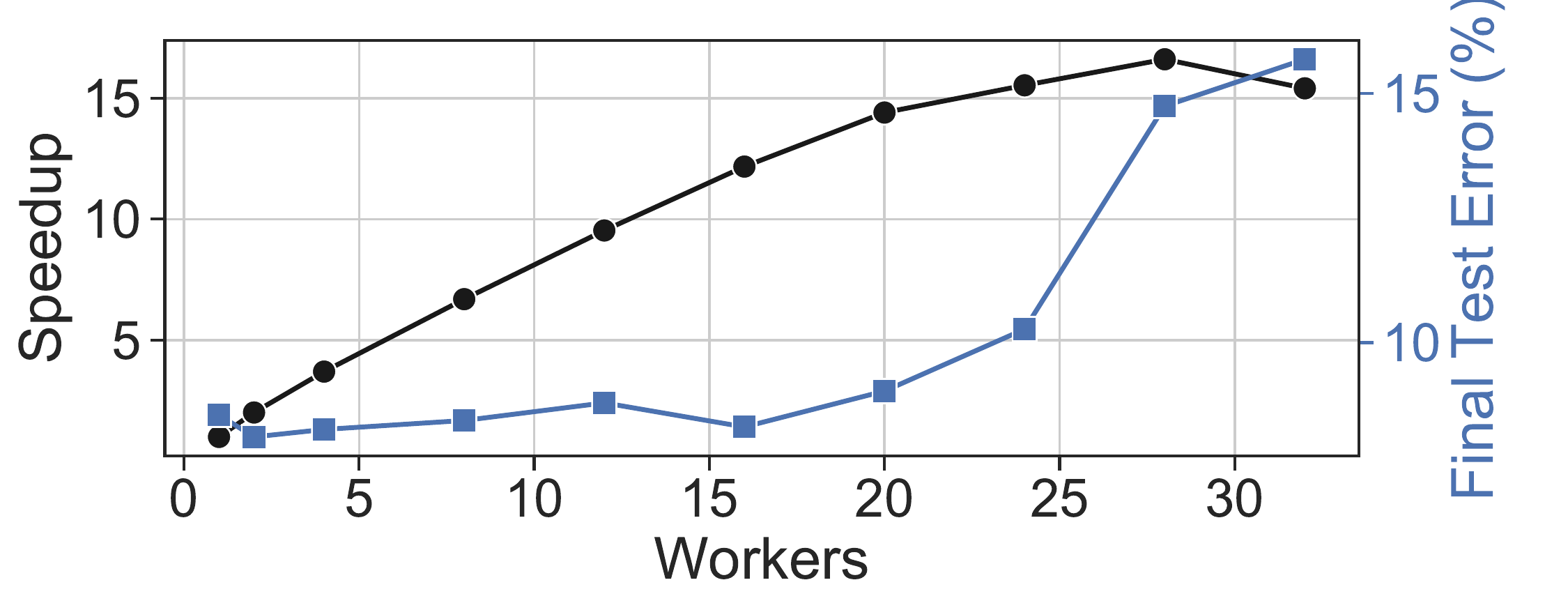}
    \caption{Speedup (solid line) and final test error (dashed) of DANA-Slim on CIFAR-10 using  ResNet-20 on Google cloud. Each worker is one Nvidia V100 GPU with a $10$Gb Ethernet NIC.}
    \label{fig:cifar_speedup}
\end{figure}

We evaluate the scalability of DANA-Slim on a public cloud data-center (Google cloud) with a single parameter server and one Nvidia Tesla V100 GPU per machine. Our cross-machine communications are based on CUDA-Aware MPI (OpenMPI) with a 10Gb Ethernet NIC per machine. \Cref{fig:cifar_speedup} shows that DANA-Slim successfully scales up to $20$ workers in agreement with the simulations of \Cref{fig:acc_cifar10_resnet} with high speedup and low final test error (less than $1\%$ higher than the baseline). We consider parameter server optimizations to be beyond the scope of this paper and detail popular parameter server optimization techniques that are orthogonal and compatible with DANA in \Cref{sec:asgd_ssgd_speedup}.

\section{Related Work}
\label{sec:related}
Asynchronous training causes gradient staleness, which hinders the convergence. Several approaches \citep{dai2018toward,staleness,pmlr-v80-zhou18b} proposed to mitigate gradient staleness by tuning the learning rate with regard to the \emph{lag} $\tau$. These approaches are designed for SGD without momentum and therefore do not address the massive \emph{gap} that momentum generates. \citet{begets} show that asynchronous training induces \emph{implicit momentum}. Thus, the momentum coefficient $\gamma$ must be decayed when scaling up the cluster size. By decreasing the \emph{gap} caused by momentum, we show that fast convergence and high final test accuracy is possible in an asynchronous environment with DANA, even when $\gamma$ is relatively high.

Other approaches for mitigating gradient staleness include DC-ASGD~\citep{dcasgd}, which uses a Taylor expansion to reduce the gradient staleness. YellowFin~\citep{mlsys2019_153} is an SGD based algorithm that automatically tunes the momentum coefficient $\gamma$ and learning rate $\eta$ throughout the training process. Both YellowFin and DC-ASGD achieve high accuracy on small clusters, but fall short when trained on large clusters (\Cref{fig:acc_cifar}) due to the massive negative effects of the gradient staleness. Communication-efficient asynchronous algorithms, such as Elastic Averaging SGD (EASGD) by \citet{elastic}, can reduce communication overhead. EASGD is an asynchronous algorithm that uses a \emph{center force} to pull the workers' parameters towards the master's parameters. This allows each worker to train asynchronously and synchronize with the master once every few iterations. Not only are these approaches compatible with (and indeed orthogonal to) DANA, we show that DANA can amplify the effectiveness of the other approaches, as demonstrated with DANA-DC (\Cref{sec:dc}).

\section{Conclusion}
In this paper we tackle gradient staleness, one of the main difficulties in scaling SGD to large clusters in an asynchronous environment. We introduce DANA: a novel asynchronous distributed technique that mitigates gradient staleness by computing the gradient on an estimated future position of the model's parameters. Despite using momentum, DANA efficiently scales to large clusters while maintaining high final accuracy and fast convergence. Thereby, we showed for the first time that momentum can be fully incorporated in asynchronous training with almost no ramifications to final accuracy. We performed an extensive set of evaluations on simulations, private compute clusters, and public cloud data-centers. Throughout our evaluations, DANA consistently outperformed all the other algorithms in both final test error and convergence rate. For future work, we plan on adapting DANA to newer optimizers, such as Nadam~\citep{nadam}, and to more recent asynchronous algorithms, in particular EASGD and YellowFin. 

\section*{Acknowledgement}
The work on this paper was supported by the Israeli Ministry of Science, Technology, and Space and by The Hasso Plattner Institute.


\bibliography{dana}
\bibliographystyle{mlsys2020}

\newpage
\appendix

\section{Experimental Setup}
\subsection{Algorithms}
\label{sec:algorithms}
\Cref{alg:NAG-ASGD_master,alg:Multi-ASGD_master,alg:DC-ASGD_master} only change the master's algorithm; the complementary worker algorithm is the same as ASGD (\Cref{alg:asgd_worker}). The master's scheme is a simple FIFO. We consider parameter server optimizations to be beyond the scope of this paper.
\begin{algorithm}[H]
\caption{NAG-ASGD: master}
\label{alg:NAG-ASGD_master}
\begin{algorithmic}
    \setlength{\itemindent}{-0.5em} 
    \STATE Receive gradient $g^i$ from worker $i$
    \STATE Update momentum $v \gets \gamma v+g^i$
    \STATE Update master's weights $\theta^0 \gets \theta^0-\eta v$
    \STATE Send $\theta^0$ to worker $i$
\end{algorithmic}
\end{algorithm}
\begin{algorithm}[H]
\caption{Multi-ASGD: master}
\label{alg:Multi-ASGD_master}
\begin{algorithmic}
    \setlength{\itemindent}{-0.5em} 
    \STATE Receive gradient $g^i$ from worker $i$
    \STATE Update momentum $v^i \gets \gamma v^i+g^i$
    \STATE Update master's weights $\theta^0 \gets \theta^0-\eta v^i$
    \STATE Send $\theta^0$ to worker $i$
\end{algorithmic}
\end{algorithm}
\begin{algorithm}[H]
\caption{DC-ASGD: master}
\label{alg:DC-ASGD_master}
\begin{algorithmic}
    \setlength{\itemindent}{-0.5em} 
    \STATE Receive gradient $g^i$ from worker $i$
    \STATE Update the gradient according to the delay compensation term $\hat{g}^i = g^i + \lambda g^i \odot g^i \odot (\theta^0 - \theta^i)$
    \STATE Update momentum $v^i \gets \gamma v^i+\hat{g}^i$
    \STATE Update master's weights $\theta^0 \gets \theta^0-\eta v^i$
    \STATE Send $\theta^0$ to worker $i$
\end{algorithmic}
\end{algorithm}
\subsection{Efficient computation in DANA-Zero} \label{sec:efficient_dana}
Computing the full summation $\sum_{i=1}^{N}v^i_{prev(i, t-1)}$ of the future position of the master's parameters can be computationally expensive since its computational cost scales up with the number of workers. In practice, DANA-Zero doesn't compute the full summation for every worker that requires the future position of the masters parameters. Instead, DANA-Zero maintains $v^0$, which is equivalent to the summation of all current momentum vectors at step $t$. After an update from worker $i$, DANA-Zero subtracts the worker's previous momentum vector $v^i_{prev(i, t-1)}$ from $v^0_t$ and adds the worker's updated momentum vector $v^i_{t}$ to $v^0_t$. Thus, after each worker update, $v^0_t$ is updated by $v^0_t = v^0_t - v^i_{prev(i, t-1)} + v^i_{t}$. Hence, the computation cost of the summation reduces from $\mathcal{O}(k \cdot N)$ to $\mathcal{O}(k)$; this is the same cost as computing the traditional NAG look-ahead, which isn't affected by the number of workers $N$.

\subsection{Datasets}
\paragraph{CIFAR} The CIFAR-10 \citep{cifar} dataset comprises of 60K RGB images partitioned into 50K training images and 10K test images. Each image contains 32x32 RGB pixels and belongs to 1 of 10 equal-sized classes. CIFAR-100 is similar but contains 100 classes. \href{https://www.cs.toronto.edu/~kriz/cifar.html}{Link}.
\paragraph{ImageNet} The ImageNet dataset~\citep{imagenet}, known as ILSVRC2012, consists of RGB images, each labeled as 1 of 1000 classes. Images are partitioned into 1.28 million training images and 50K validation images. Each image is randomly cropped and re-sized to 224x224 (1-crop validation). \href{http://www.image-net.org/}{Link}.

\subsection{Gamma Distribution}
\label{sec:gamma_distribution}
\citet{Ali:2000:TET} suggest a method called \emph{CVB} to simulate the run-time of a distributive network of computers. The method is based on several definitions:
Task execution time variables:
\begin{itemize}
    \item $\mu_{task}$ - mean time of tasks
    \item $V_{task}$ - variance of tasks
    \item $\mu_{mach}$ - mean computation power of machines
    \item $V_{mach}$ - variance of computation power of machines
    \item $\alpha_{task} = \frac{1}{V_{task}^2}$
    \item $\alpha_{mach} = \frac{1}{V_{mach}^2}$
\end{itemize}
$G(\alpha, \beta)$ is a random number generated using a gamma distribution, where $\alpha$ is the shape and $\beta$ is the scale.

For our case, all tasks are similar and run on a batch size of B. Therefore, the algorithm for deciding the execution-time of every task on a certain machine is reduced to one of the following:
\begin{algorithm}[H]
\caption{Task execution time - homogeneous machines}
\label{alg:task_exe_homo}
\begin{algorithmic}
    \setlength{\itemindent}{-0.5em} 
    \STATE $\beta_{task}$ = $\frac{\mu_{task}}{\alpha_{task}}$
    \STATE $q = G(\alpha_{task}, \beta_{task})$
	\STATE $\beta_{mach} = \frac{q}{\alpha_{mach}}$
    \STATE for i from 0 to $K-1$:
	\STATE $\quad time = G(\alpha_{mach}, \beta_{mach})$
\end{algorithmic}
\end{algorithm}
\begin{algorithm}[H]
\caption{Task execution time - heterogeneous machines}
\label{alg:task_exe_hetero}
\begin{algorithmic}
    \setlength{\itemindent}{-0.5em} 
    \STATE $\beta_{mach}$ = $\frac{\mu_{mach}}{\alpha_{mach}}$
    \STATE for j from 0 to $M$:
    \STATE $\quad p[j] = G(\alpha_{mach}, \beta_{mach})$ 
	\STATE $\beta_{task}[j] = \frac{p[j]}{\alpha_{task}}$
    \STATE for i from 0 to $K-1$:
	\STATE $\quad time = G(\alpha_{task}, \beta_{task}[curr])$ 
\end{algorithmic}
\end{algorithm}
where $K$ is the total amount of tasks of all the machines combined (the total number of batch iterations), $M$ is the total number of machines (workers), and $curr$ is the machine currently about to run.

We note that \Cref{alg:task_exe_homo,alg:task_exe_hetero} naturally give rise to stragglers. In the homogeneous algorithm, all workers have the same mean execution time but some tasks can still be very slow; this generally means that in every epoch a different machine will be the slowest. In the heterogeneous algorithm, every machine has a different mean execution time throughout the training. We further note that $p[j]$ is the mean execution time of machine $j$ on the average task. 

In our experiments, we simulated execution times using the following parameters as suggested by \citet{Ali:2000:TET}: $\mu_{task} = \mu_{mach} = B\cdot V_{mach}^2$, where $B$ is the batch size, yielding a mean execution time of $\mu$ simulated time units, which is proportionate to $B$. In the homogeneous setting $V_{mach} = 0.1$, whereas in the heterogeneous setting $V_{mach} = 0.6$. For both settings, $V_{task} = 0.1$.

\Cref{fig:gamma_distribution} illustrates the differences between the homogeneous and heterogeneous gamma-distribution. Both environments have the same mean (128) but the probability of having an iteration that is at least 1.25x longer than the mean (which means 160 or more) is significantly higher in the heterogeneous environment (27.9\% in heterogeneous environment as opposed to 1\% in the homogeneous environment).

\subsection{Hyperparameters}
\label{sec:hyperparameters}
To verify that decreasing the \emph{gap} leads to a better final test error and convergence rate, especially when scaling to more workers, we used the same hyperparameters across all the tested algorithms. These hyperparameters are the original hyperparameters of the respective neural network architecture, which are tuned for a single worker. 

\paragraph{CIFAR-10 ResNet-20}
\begin{itemize}
    \item Initial learning rate $\eta$: $0.1$
    \item Momentum coefficient $\gamma$: $0.9$ with NAG
    \item Dampening: $0$ (no dampening)
    \item Batch size $B$: $128$
    \item Weight decay: $1e-4$
    \item Learning rate decay: $0.1$
    \item Learning rate decay schedule: epochs $80$ and $120$
    \item Total epochs: $160$
\end{itemize}

\paragraph{CIFAR-10/100 Wide ResNet 16-4}
\begin{itemize}
    \item Initial learning rate $\eta$: $0.1$
    \item Momentum coefficient $\gamma$: $0.9$ with NAG
    \item Dampening: $0$ (no dampening)
    \item Batch size $B$: $128$
    \item Weight decay: $5e-4$
    \item Learning rate decay: $0.2$
    \item Learning rate decay schedule: epochs $60$, $120$ and $160$
    \item Total epochs: $200$
\end{itemize}

\paragraph{ImageNet ResNet-50}
\begin{itemize}
    \item Initial learning rate $\eta$: $0.1$
    \item Momentum coefficient $\gamma$: $0.9$ with NAG
    \item Dampening: $0$ (no dampening)
    \item Batch size $B$: $256$
    \item Weight decay: $1e-4$
    \item Learning rate decay: $0.1$
    \item Learning rate decay schedule: epochs $30$ and $60$
    \item Total epochs: $90$
\end{itemize}

\paragraph{Learning Rate Warm-Up} In the early stages of training, the network generally changes rapidly, causing error spikes. For all algorithms, we followed the gradual warm-up approach proposed by~\citet{goyal2017accurate} to overcome this problem. We divided the initial learning rate by the number of workers $N$ and ramped it up linearly until it reached its original value after five epochs. We also used momentum correction \citep{goyal2017accurate} in all algorithms to stabilize training when the learning rate changes.

\section{Additional Results}
\subsection{CIFAR Final Accuracies}
\label{sec:cifar_final}
DANA-DC starts to show signs of divergence when it reaches  $32$. However, when we tuned the learning rate $\eta$, DANA-DC reached a significantly lower test error than that shown in \Cref{tab:resnet-20-table,tab:wr-cifar10-table,tab:wr-cifar100-table}. More precisely, when trained on $32$ workers with $\eta = 0.025$, DANA-DC reached a test error of only 2.5\% higher than the baseline on both CIFAR10 scenarios and 4.5\% higher than the baseline on CIFAR100.

\begin{table*}[t]
\label{sec:cifar_final_accuracy}
\captionof{table}{ResNet-20 CIFAR10 Final Test Accuracy (Baseline 91.63\%)}
\label{tab:resnet-20-table}
\centering
\begin{tabular}{c c c c c c c}
\toprule
\#Workers &         DANA-DC &       DANA-Slim &         DC-ASGD &       Multi-ASGD &         NAG-ASGD &       YellowFin \\
\midrule
4        &  \textbf{91.79 $\pm$ 0.21} &  91.65 $\pm$ 0.15 &  91.68 $\pm$ 0.18 &   91.55 $\pm$ 0.15 &   91.41 $\pm$ 0.23 &   90.05 $\pm$ 0.37 \\
\midrule
8        &  91.51 $\pm$ 0.16 &  \textbf{91.52 $\pm$ 0.16} &  90.67 $\pm$ 0.26 &   91.28 $\pm$ 0.21 &   90.83 $\pm$ 0.18 &   90.29 $\pm$ 0.14 \\
\midrule
12       &  \textbf{91.49 $\pm$ 0.18} &  91.32 $\pm$ 0.16 &  72.16 $\pm$ 5.32 &   90.42 $\pm$ 0.08 &   82.41 $\pm$ 4.41 &  90.54 $\pm$ 0.18 \\
\midrule
16       &  91.01 $\pm$ 0.25 &  \textbf{91.02 $\pm$ 0.16} &  18.35 $\pm$ 16.7 &   84.88 $\pm$ 1.28 &  17.45 $\pm$ 12.39 &  41.19 $\pm$ 38.21 \\
\midrule
20       &  \textbf{90.78 $\pm$ 0.32} &  90.56 $\pm$ 0.32 &    10.0 $\pm$ 0.0 &   57.32 $\pm$ 23.9 &   10.17 $\pm$ 0.33 &     10.0 $\pm$ 0.0 \\
\midrule
24       &  89.76 $\pm$ 0.37 &   \textbf{89.81 $\pm$ 0.4} &  17.65 $\pm$ 15.3 &  23.83 $\pm$ 18.43 &   19.81 $\pm$ 12.2 &     10.0 $\pm$ 0.0 \\
\midrule
28       &  \textbf{87.82 $\pm$ 0.83} &   84.91 $\pm$ 3.6 &    10.0 $\pm$ 0.0 &   15.58 $\pm$ 9.38 &    18.5 $\pm$ 17.0 &     10.0 $\pm$ 0.0 \\
\midrule
32       &  \textbf{82.99 $\pm$ 3.38} &  79.33 $\pm$ 4.68 &    10.0 $\pm$ 0.0 &   12.04 $\pm$ 4.08 &   12.06 $\pm$ 4.11 &     10.0 $\pm$ 0.0 \\
\bottomrule
\end{tabular}
\end{table*}

\begin{table*}[t]
\captionof{table}{Wide ResNet 16-4 CIFAR10 Final Test Accuracy (Baseline $95.17\%$)}
\label{tab:wr-cifar10-table}
\centering
\begin{tabular}{c c c c c c c}
\toprule
\#Workers &          DANA-DC &        DANA-Slim &          DC-ASGD &       Multi-ASGD &         NAG-ASGD &       YellowFin \\
\midrule
4        &  \textbf{95.08 $\pm$ 0.13} &  95.04 $\pm$ 0.11 &   93.66 $\pm$ 0.16 &   94.97 $\pm$ 0.11 &   94.81 $\pm$ 0.11 &   92.81 $\pm$ 0.04 \\
\midrule
8        &  94.84 $\pm$ 0.19 &   \textbf{94.91 $\pm$ 0.2} &    89.76 $\pm$ 1.0 &   94.25 $\pm$ 0.12 &   92.83 $\pm$ 0.61 &   93.52 $\pm$ 0.08 \\
\midrule
12       &  94.35 $\pm$ 0.16 &  \textbf{94.45 $\pm$ 0.21} &  18.24 $\pm$ 16.49 &   92.64 $\pm$ 0.17 &  44.35 $\pm$ 28.22 &   93.78 $\pm$ 0.08 \\
\midrule
16       &  \textbf{93.67 $\pm$ 0.15} &   93.66 $\pm$ 0.2 &  22.29 $\pm$ 24.59 &  70.86 $\pm$ 14.08 &  23.36 $\pm$ 26.72 &   93.19 $\pm$ 0.18 \\
\midrule
20       &  \textbf{92.73 $\pm$ 0.37} &  92.72 $\pm$ 0.13 &  24.59 $\pm$ 29.18 &  16.13 $\pm$ 12.25 &  33.41 $\pm$ 30.31 &  26.03 $\pm$ 32.06 \\
\midrule
24       &    \textbf{90.4 $\pm$ 0.4} &  88.76 $\pm$ 1.85 &  24.27 $\pm$ 28.54 &   26.98 $\pm$ 23.0 &   11.62 $\pm$ 3.23 &     10.0 $\pm$ 0.0 \\
\midrule
28       &  \textbf{76.88 $\pm$ 4.65} &   71.62 $\pm$ 5.0 &  34.92 $\pm$ 30.88 &  31.77 $\pm$ 26.66 &  31.91 $\pm$ 17.68 &     10.0 $\pm$ 0.0 \\
\midrule
32       &  \textbf{69.35 $\pm$ 6.86} &  69.13 $\pm$ 6.85 &     10.0 $\pm$ 0.0 &    14.51 $\pm$ 7.4 &  19.52 $\pm$ 19.03 &     10.0 $\pm$ 0.0 \\
\bottomrule
\end{tabular}
\end{table*}

\begin{table*}[t]
\captionof{table}{Wide ResNet 16-4 CIFAR100 Final Test Accuracy (Baseline $76.72\%$)}
\label{tab:wr-cifar100-table}
\centering
\begin{tabular}{c c c c c c c}
\toprule
\#Workers &          DANA-DC &        DANA-Slim &          DC-ASGD &       Multi-ASGD &         NAG-ASGD &       YellowFin \\
\midrule
4        &  76.22 $\pm$ 0.15 &  \textbf{76.27 $\pm$ 0.33} &  74.03 $\pm$ 0.26 &  76.07 $\pm$ 0.23 &   76.27 $\pm$ 0.2 &   66.91 $\pm$ 0.25 \\
\midrule
8        &  76.05 $\pm$ 0.23 &  \textbf{76.07 $\pm$ 0.17} &  70.48 $\pm$ 0.48 &  75.33 $\pm$ 0.29 &  74.24 $\pm$ 0.27 &   68.05 $\pm$ 0.21 \\
\midrule
12       &  75.57 $\pm$ 0.24 &  \textbf{75.64 $\pm$ 0.28} &   65.7 $\pm$ 0.68 &  73.63 $\pm$ 0.29 &  69.29 $\pm$ 0.56 &   69.36 $\pm$ 0.22 \\
\midrule
16       &  74.69 $\pm$ 0.28 &   \textbf{74.97 $\pm$ 0.1} &   56.5 $\pm$ 1.75 &  70.68 $\pm$ 0.23 &  67.37 $\pm$ 0.74 &   69.85 $\pm$ 0.29 \\
\midrule
20       &  73.14 $\pm$ 0.58 &  \textbf{73.48 $\pm$ 0.35} &  45.61 $\pm$ 4.16 &  68.12 $\pm$ 0.48 &  37.98 $\pm$ 7.21 &   69.62 $\pm$ 0.18 \\
\midrule
24       &  71.19 $\pm$ 0.32 &  \textbf{71.91 $\pm$ 0.27} &  50.24 $\pm$ 2.63 &  66.12 $\pm$ 0.58 &   9.67 $\pm$ 4.89 &    67.9 $\pm$ 0.49 \\
\midrule
28       &  69.14 $\pm$ 0.62 &  \textbf{69.77 $\pm$ 0.83} &  48.49 $\pm$ 1.31 &   54.3 $\pm$ 2.29 &   6.35 $\pm$ 7.41 &  13.76 $\pm$ 25.52 \\
\midrule
32       &  67.19 $\pm$ 0.79 &   \textbf{67.91 $\pm$ 0.7} &  45.98 $\pm$ 2.26 &  29.43 $\pm$ 8.11 &  12.71 $\pm$ 7.69 &      1.0 $\pm$ 0.0 \\
\bottomrule
\end{tabular}
\end{table*}

\subsection{ImageNet Final Accuracies}
\label{sec:imagenet_final_accuracy}
\begin{table*}[t]
\captionof{table}{ResNet-50 ImageNet Final Test Accuracy (Baseline $75.64\%$)}
\label{tab:imagenet-table}
\centering
\begin{tabular}{c c c c c c c c}
\toprule
\#Workers & DANA-DC & DANA-Slim & DC-ASGD & Multi-ASGD & NAG-ASGD & YellowFin &  LWP \\
\midrule
16 &  \textbf{75.54\%} & 74.95\% &  72.64\% &     74.96\% &   73.22\% &    53.74\% & 74.12\%\\
\midrule
32 & 74.86\% & \textbf{74.89\%} &  59.99\% &     71.72\% &   70.64\% &    57.88\% & 71.84\%\\
\midrule
48 &   \textbf{73.80\%} &   73.75\% &  31.71\% &     65.13\% &   66.78\% &    63.07\% & 67.34\%\\
\midrule
64  &  \textbf{73.58\%} &    69.88\% &    8.1\% &     54.04\% &   59.81\% &     0.15\% & 61.8\%\\
\midrule
128 &   \textbf{69.50\%} &       NaN &     NaN &        NaN &      NaN &       NaN & NaN\\
\bottomrule
\end{tabular}
\end{table*}

\Cref{tab:imagenet-table} lists the final test accuracy of the different algorithms when training the ResNet-50 architecture \citep{resnet} on ImageNet. DANA consistently outperforms all the other algorithms.

\subsection{Normalized Gap}
\label{sec:normalized_gap}

\begin{figure*}[t]
\centering
    \begin{subfigure}{0.49\textwidth}
            \includegraphics[width=\columnwidth]{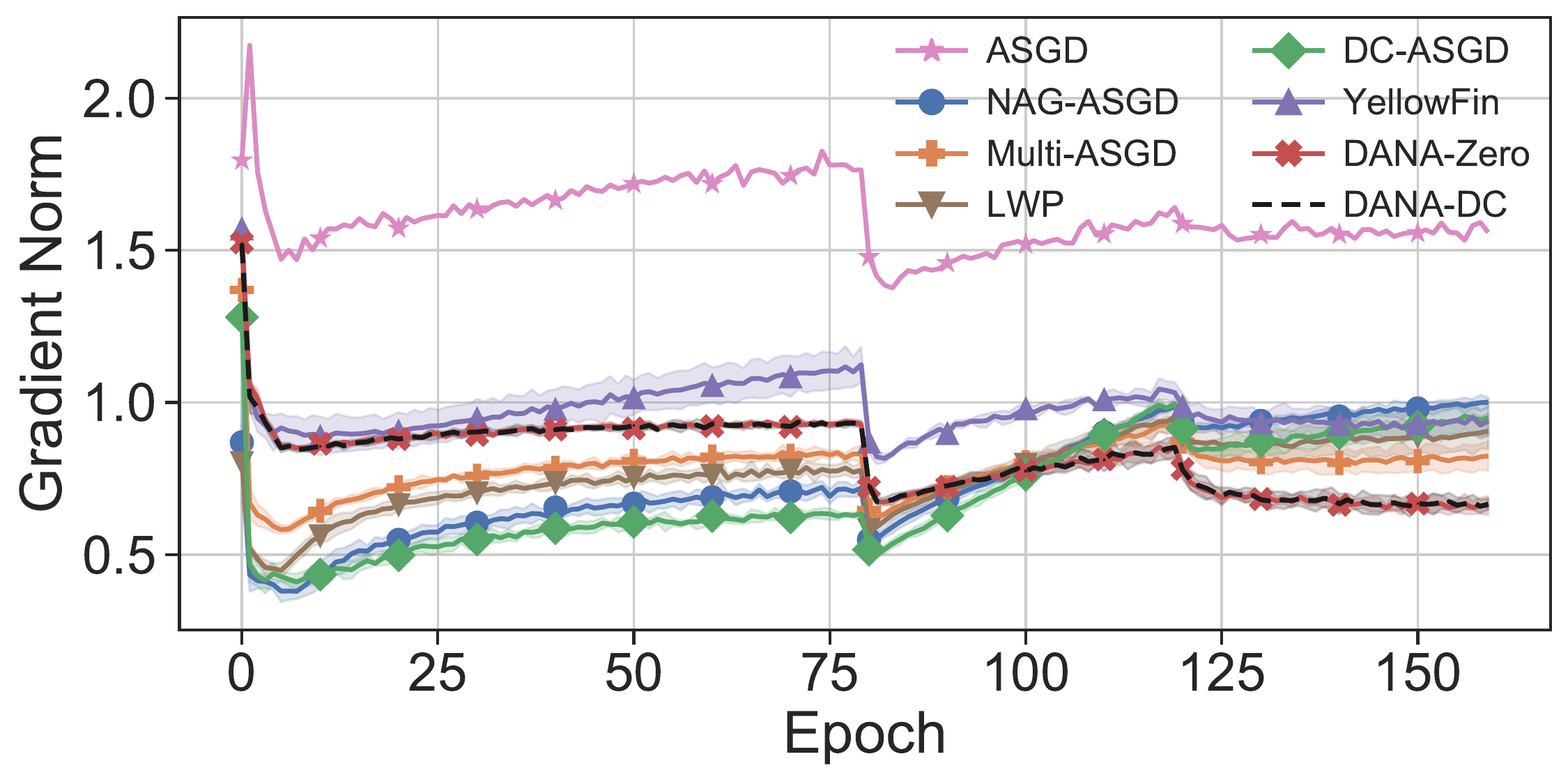}
        	\caption{Gradient norm.}
        	\label{fig:gradient_norm}
    \end{subfigure}
    \begin{subfigure}{0.49\textwidth}
            \includegraphics[width=\columnwidth]{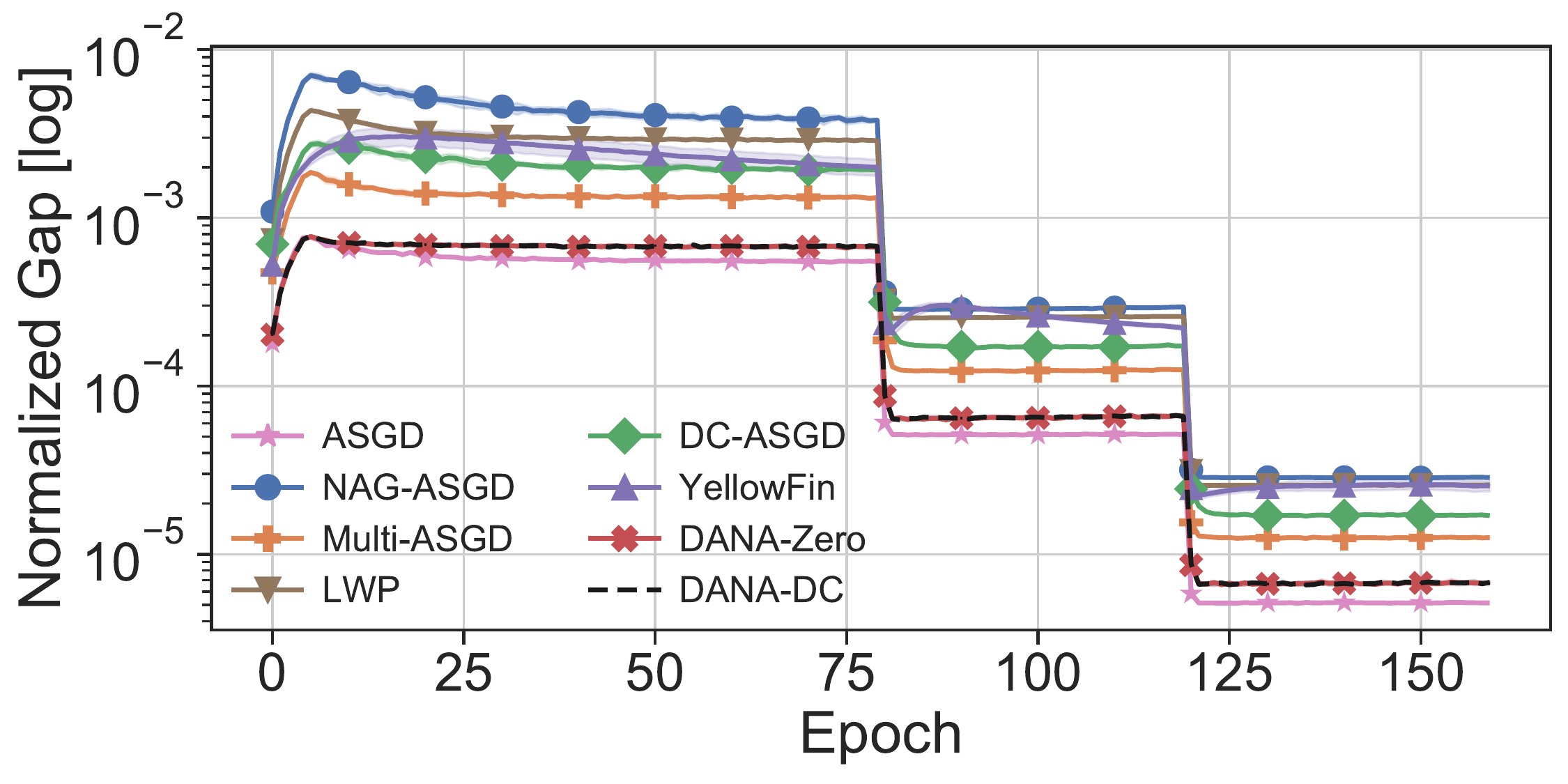}
        	\caption{Normalized gap.}
        	\label{fig:normalized_gap}
    \end{subfigure}
    \caption{\Cref{fig:gradient_norm,fig:normalized_gap} compare the different asynchronous algorithms when training the ResNet-20 architecture on the CIFAR-10 dataset with $8$ workers. \Cref{fig:gradient_norm,fig:normalized_gap} show the gradient norm and the \emph{normalized gap}, respectively, throughout the training process. The large drops in \Cref{fig:normalized_gap} are caused by learning rate decay.}
\label{fig:norms}
\end{figure*}

As shown in \Cref{fig:rmse_algos}, the \emph{gap} of DANA-Zero is smaller than that of ASGD, despite \Cref{thm:dana_gap}. This is because DANA-Zero uses momentum, which accelerates the convergence and leads to smaller gradients. To make a more accurate comparison between the gaps of different algorithms, we define the \emph{normalized gap} as $\GAP^*(\Delta_{t+\tau})=\frac{\GAP(\Delta_{t+\tau})}{\norm{g_t}}$.
\Cref{fig:normalized_gap} shows that the \emph{normalized gap} of ASGD is roughly similar to that of DANA-Zero, empirically confirming \Cref{thm:dana_gap}. This suggests that DANA-Zero's future position approximation is close to optimal, even when training with stragglers.

\section{Asynchronous Speedup}
\label{sec:asgd_ssgd_speedup}

\subsection{CIFAR-10 Distributed Experiments}
\label{sec:cifar_distributed}

DANA-Slim successfully scales up to $20$ workers (similarly to the simulations) with high speedup and low final test error (less than $1\%$ higher than the baseline). Above $20$ workers, the master becomes a bottleneck. This is consistent with the literature on ASGD \citep{xing2015petuum}. To overcome this bottleneck, existing parameter server optimization techniques can be used, such as sharding \citep{downpour,li2014scaling}, synchronization frequency reduction \citep{wang2018adaptive}, network utilization improvements \citep{jia2018beyond,mlsys2020_173,mlsys2020_81}, communication compression \citep{tang2019doublesqueeze,lim20183lc,lin2018deep,signsgd,mlsys2019_32}, and overlapping communications with computations \citep{mlsys2019_199,mlsys2019_75}. We note that DANA is compatible with these optimizations. The speedup superiority of ASGD methods compared to SSGD methods is discussed in \Cref{sec:asgd_ssgd_speedup}. 

\Cref{fig:cifar_speedup} shows the speedup and final test error when running DANA-Slim on the Google Cloud Platform with a single parameter server (master) and one Nvidia Tesla V100 GPU per machine, when training ResNet-20 on the CIFAR-10 dataset. It shows speedup of up to $16\times $ when training with $N=24$ workers. As before, its final test error remains close to the baseline for up to $N=24$ workers.

\begin{figure*}[t]
\centering
    \begin{subfigure}{0.48\textwidth}
            \includegraphics[width=\columnwidth]{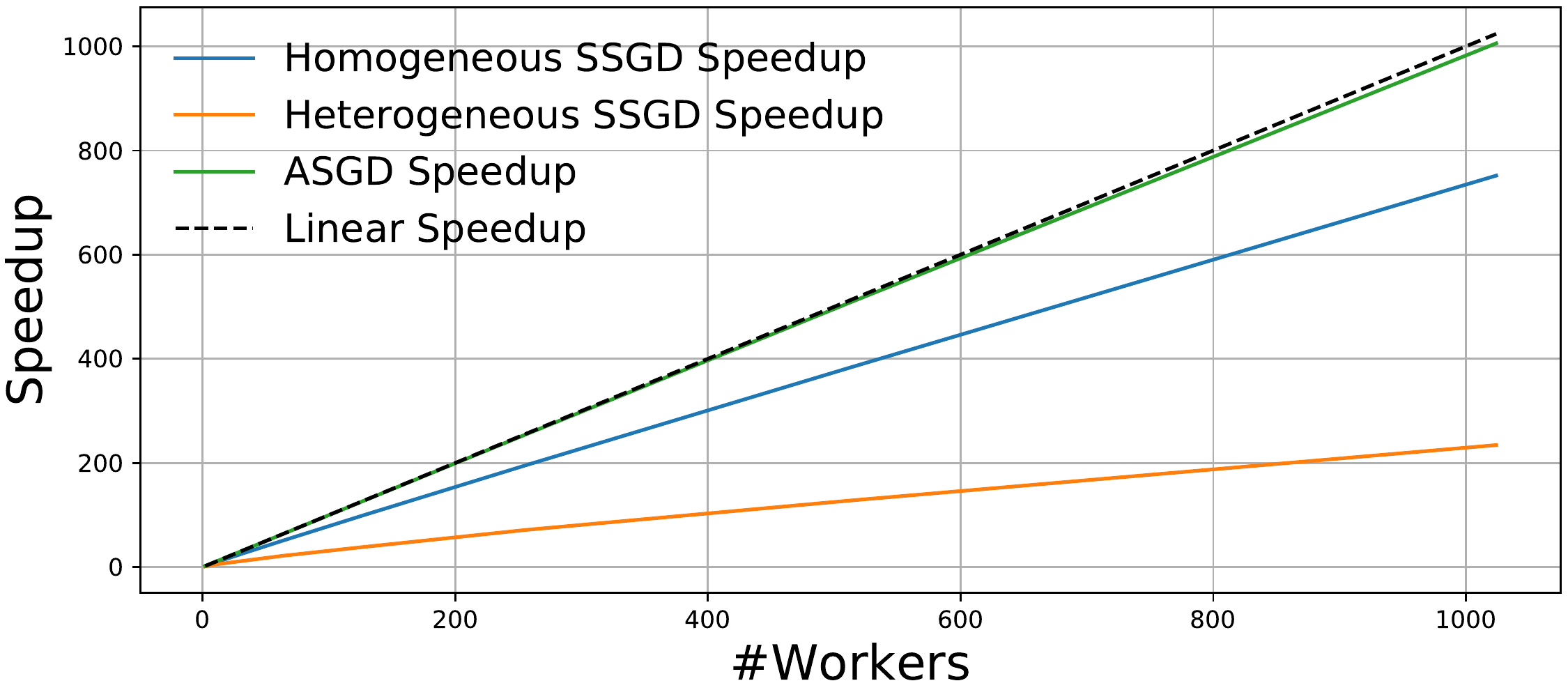}
        	\caption{ASGD and SSGD speedups.}
        	\label{fig:GammaSpeedup}
    \end{subfigure}
    \begin{subfigure}{0.48\textwidth}
            \includegraphics[width=\columnwidth]{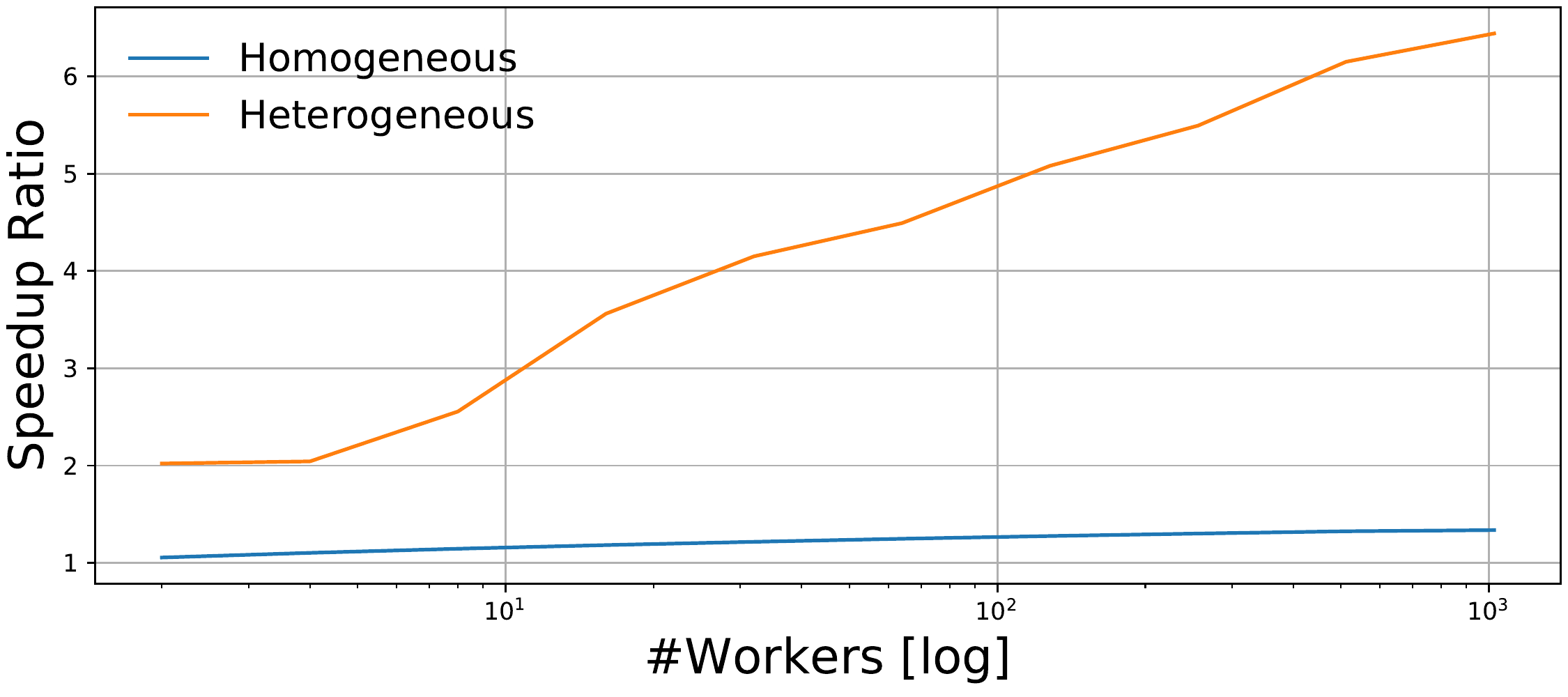}
        	\caption{ASGD speedup over SSGD. X axis is in log scale.}
        	\label{fig:GammaSpeedupRatio}
    \end{subfigure}
    \caption{Theoretical speedups for any ASGD (such as DANA) and SSGD algorithms when batch execution times are drawn from a gamma distribution. Communication overheads aren't modeled. Since ASGD is more efficient in communication, accounting for the communication overheads should expand the gap between the ASGD and SSGD, in favor of ASGD.}
\label{fig:gamma-speedup}
\end{figure*}

Cloud computing is becoming increasingly popular as a platform to perform distributed training of deep neural networks \citep{mlsys2020_168}. Although synchronous SGD is currently the primary method \citep{fast_imagenet_1,fast_imagenet_2,fast_imagenet_3,goyal2017accurate,mlsys2020_33} used to distribute the learning process, it suffers from substantial slowdowns when run in non-dedicated environments such as the cloud. This shortcoming is magnified in heterogeneous environments and non-dedicated networks. ASGD addresses the SSGD drawback and enjoys linear speedup in terms of the number of workers in both heterogeneous and homogeneous environments even in non-dedicated networks. This makes ASGD a potentially better alternative for cloud computing. 

\Cref{fig:GammaSpeedup} shows the theoretically achievable speedup, based on the detailed gamma-distributed model, for asynchronous (GA and other ASGD variants) and synchronous algorithms using homogeneous and heterogeneous workers. The asynchronous algorithms can achieve linear speedup while the synchronous algorithm (SSGD) falls short as we increase the number of workers. This occurs because SSGD must wait in each iteration until all workers complete their batch. \Cref{fig:GammaSpeedupRatio} shows that ASGD-based algorithms (including GA, SA and DANA versions) are up to $21\%$ faster than SSGD in homogeneous environments. In heterogeneous environments, ASGD methods can be 6x faster than SSGD. We note that this speedup is an underestimate, since our simulation includes only batch execution times. It does not model the execution time of barriers, all-gather operations, and other overheads which usually increase communication time, especially in SSGD.

\section{Heterogeneous Environment}
\label{sec:hetero}
\begin{figure*}[t]
\centering
    \begin{subfigure}{0.49\textwidth}
            \includegraphics[width=\columnwidth]{hetero_acc.pdf}
        	\caption{Final test error for different numbers of workers $N$.}
        	\label{fig:hetero_acc}
    \end{subfigure}
    \begin{subfigure}{0.49\textwidth}
            \includegraphics[width=\columnwidth]{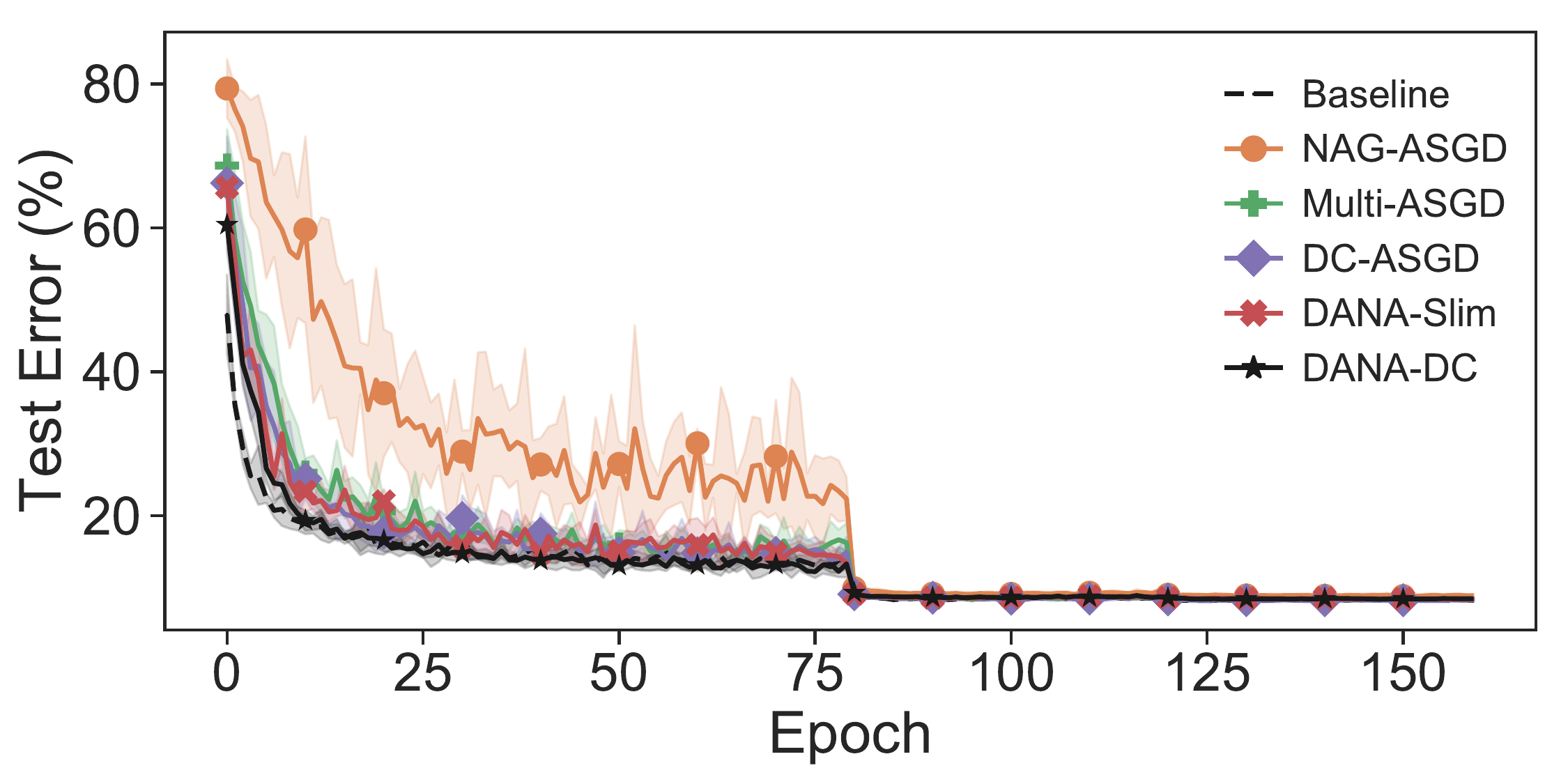}
        	\caption{Convergence rate for 8 workers.}
        	\label{fig:hetero_conv}
    \end{subfigure}
    \caption{Training of ResNet-20 on CIFAR10 in a heterogeneous environment.}
\label{fig:hetero}
\end{figure*}
In our experiments, the algorithms scale better in the heterogeneous environment \Cref{fig:hetero_acc} than in the homogeneous environment (\Cref{fig:acc_cifar10_resnet}). The reason is that stragglers naturally have less impact on the training process. We will demonstrate this with a simplistic example. Consider an asynchronous environment with only $N=2$ workers, where one worker is significantly faster than the other. The fast worker will run as in sequential SGD, since its \emph{gap} and \emph{lag} will mostly be zero. Conversely, the slow workers will have minimal impact and therefore its stale gradients wouldn't harm the convergence process.

This suggests that high accuracy can be attained more easily in asynchronous heterogeneous environments than in homogeneous environments. \Cref{fig:hetero_acc,fig:hetero_conv} show that even in a heterogeneous environment, DANA-DC converges the fastest and achieves the highest final accuracy. The final accuracies are listed in \Cref{tab:hetero-resnet-cifar10-table} below.

\begin{table*}
\captionof{table}{Heterogeneous Environment ResNet 20 CIFAR10 Final Test Accuracy (Baseline 91.63\%)}
\label{tab:hetero-resnet-cifar10-table}
\centering
\begin{tabular}{c c c c c c}
\toprule
\#Workers &          DANA-DC &        DANA-Slim &          DC-ASGD &       Multi-ASGD &         NAG-ASGD \\
\midrule
4        &  91.57 $\pm$ 0.14 &   91.7 $\pm$ 0.18 &    91.6 $\pm$ 0.14 &  \textbf{91.77 $\pm$ 0.22} &   91.38 $\pm$ 0.12 \\
\midrule
8        &  91.57 $\pm$ 0.18 &  91.55 $\pm$ 0.28 &   \textbf{91.72 $\pm$ 0.21} &  91.59 $\pm$ 0.11 &   91.15 $\pm$ 0.23 \\
\midrule
16       &  91.28 $\pm$ 0.21 & \textbf{ 91.31 $\pm$ 0.17} &    90.98 $\pm$ 0.5 &   91.12 $\pm$ 0.3 &   83.65 $\pm$ 5.17 \\
\midrule
24       &  \textbf{91.21 $\pm$ 0.19} &  90.94 $\pm$ 0.27 &   90.11 $\pm$ 0.92 &   89.6 $\pm$ 2.03 &  39.36 $\pm$ 36.01 \\
\midrule
32       &  90.33 $\pm$ 0.58 &  \textbf{90.52 $\pm$ 1.04} &  57.62 $\pm$ 38.93 &  74.18 $\pm$ 32.1 &  37.52 $\pm$ 34.12 \\
\bottomrule
\end{tabular}
\end{table*}

\end{document}